\documentclass{article} 
\usepackage{hyperref}
\usepackage{supertabular}
\usepackage{adjustbox}
\usepackage{tabularray}
\usepackage{graphicx}
\usepackage{multicol}  
\usepackage[numbers]{natbib}  
\usepackage{float}
\usepackage{amsmath}
\usepackage{booktabs} 
\usepackage{caption} 
\usepackage{longtable}

\title{Benchmarking Classical, Deep, and Generative Models for Human Activity Recognition}

\author{Md Meem Hossain\textsuperscript{1*}, The Anh Han\textsuperscript{1}, Safina Showkat Ara\textsuperscript{2}and Zia Ush Shamszaman\textsuperscript{1*}\\[2ex]
\textsuperscript{1}Department of Computing and Games, Teesside University, Middlesbrough, TS1 3BX, United Kingdom.\\[2ex]
*Corresponding author(s). E-mail(s): md.hossain@tees.ac.uk; z.shamszaman@tees.ac.uk\\
Contributing authors: md.hossain@tees.ac.uk; t.han@tees.ac.uk; safina.ara@sunderland.ac.uk; z.shamszaman@tees.ac.uk
}

\makeatletter
\renewcommand{\@maketitle}{%
  \newpage
  \null
  \vskip 2em%
  \begin{center}%
    {\LARGE \@title \par}%
    \vskip 1.5em%
    {\large
      \@author \par}%
    \vskip 1em%
  \end{center}%
  \par
  \vskip 1.5em}
\makeatother

\begin{document}

\maketitle 
\begin{abstract}
Human Activity Recognition (HAR) has gained significant importance with the growing use of sensor-equipped devices and large datasets. This paper evaluates the performance of three categories of models : classical machine learning, deep learning architectures, and Restricted Boltzmann Machines (RBMs) using five key benchmark datasets of HAR (UCI-HAR, OPPORTUNITY, PAMAP2, WISDM, and Berkeley MHAD). We assess various models, including Decision Trees, Random Forests, Convolutional Neural Networks (CNN), and Deep Belief Networks (DBNs), using metrics such as accuracy, precision, recall, and F1-score for a comprehensive comparison. The results show that CNN models offer superior performance across all datasets, especially on the Berkeley MHAD. Classical models like Random Forest do well on smaller datasets but face challenges with larger, more complex data. RBM-based models also show notable potential, particularly for feature learning. This paper offers a detailed comparison to help researchers choose the most suitable model for HAR tasks. 
\end{abstract}
\noindent\textbf{Keywords:} Human Activity Recognition, Machine Learning, Deep Learning, Restricted Boltzmann Machines, Performance Analysis
\section{Introduction}
In recent years, Human Activity Recognition (HAR) has become increasingly important due to its wide range of applications \citep{wang2019deep} in healthcare \citep{zhou2020deep}, smart homes\citep{bouchabou2021survey}, security \citep{abdel2020st}, and human-computer interaction \citep{gammulle2023continuous}. The rapid development of sensor technology has further enhanced the potential of HAR to revolutionize these fields. While HAR research has been conducted since the late 1990s  \citep{xu2021human}, technological advancements have accelerated progress and expanded the possibilities for accurately recognizing human activities. Sensor-based and vision-based methods are two main approaches for collecting data for HAR \citep{dang2020sensor}. Vision-based HAR involves using visual information from surveillance cameras, smartphones and wearable cameras or special cameras \citep{zhang2017review}. Sensors-based approaches rely on data collected from various sensors such as GPS, accelerometers, gyroscopes, microphones, magnetometers, and inertial measurement units (IMUs) \citep{li2018comparison}. These sensors are commonly embedded in smartphones \citep{shoaib2015survey}, smartwatches\citep{wang2019survey}, fitness trackers and other available devices \citep{liu2016action}. Over the past decade, HAR has benefited from advancements in sensor technology \citep{antar2019challenges}, especially the development of low-power, low-cost, high-capacity, miniaturized sensors, and wire and wireless communication networks \citep{cook2013transfer,dang2020sensor,ding2011sensor,guerrero2018sensor}. As sensor technology has evolved to provide more sophisticated and diverse data, researchers are now working on integrating data from multiple sensors and sources \citep{chen2012sensor}. This integration enables context processing \citep{dey2001understanding}, developing advanced algorithms and technology for activity recognition and inference \citep{demrozi2021b}, and developing more complex and practical HAR applications. Machine Learning \citep{shekhar2023human,tsinganos2018comparison,ramasamy2018recent,monir2021machine}, Deep Learning \citep{abdel2020st,zhou2020deep,alhumayyani2021deep,hammerla2016deep,wang2019deep,monir2021machine,tee2022close,gupta2021deep} and other Artificial Intelligence approaches play a vital role in extracting meaningful patterns and features to recognize and classify the precise activity or behaviour by an individual at a specific instant utilizing sensor data \citep{sousa2019human}.

Despite the significant advancements in HAR, it remains a challenging task due to the diversity and complexity of human activities. One of the primary challenges lies in developing models that can effectively and efficiently recognize various activities across different scenarios. The complexity of activity patterns, the diversity of data sources (e.g., wearable sensors, video data), and the growing size of datasets make it difficult to design a single algorithm \citep{arshad2022human} that generalizes well across multiple contexts. Researchers have explored a broad spectrum of techniques, from traditional machine learning methods  \citep{ramasamy2018recent,hou2020study} to sophisticated deep learning architectures \citep{plotz2018deep,hammerla2016deep,gupta2021deep}. Each of these approaches has its unique strengths and limitations, and understanding their comparative performance  \citep{wang2016comparative} is crucial for making informed decisions in real-world applications \citep{islam2022human}

While many models have been developed for HAR, there is a pressing need for a comprehensive evaluation across different model families, including classical machine learning, deep learning, and generative models. Such a comparison is crucial for understanding which models perform best under varying conditions and data complexities, guiding practitioners in selecting the right approach for their specific needs. Previous works  have shown remarkable progress in the development of sophisticated algorithms and approaches for extracting useful insights from HAR data \citep{vrigkas2015review,jobanputra2019human,weiss2016investigation}. Researchers have investigated a variety of HAR techniques, including classic machine learning methods and more advanced deep learning architectures \citep{shekhar2023human,mauldin2018smartfall,wang2019deep,ramasamy2018recent,tee2022close,gupta2021deep}. Traditional machine learning algorithms, such as Support Vector Machines (SVMs), K-nearest Neighbour (KNN), and Decision Trees, have been shown highly capable of accurately recognizing human behaviours \citep{alhumayyani2021deep,mauldin2018smartfall}. These classic models have proven effective in understanding diverse activity patterns and providing useful insights into a variety of settings \citep{alhumayyani2021deep}. However, these models may not be able to capture the complex temporal patterns of human activities \citep{weiss2016investigation}. To address these limitations, researchers have shifted to deep learning architectures like Convolutional Neural Networks (CNN), Recurrent Neural Networks (RNN), and other Artificial Neural Networks (ANN) architectures for enhancing  accuracy on a wide range of HAR tasks \citep{hammerla2016deep,mauldin2018smartfall,wang2019deep,li2018comparison,caya2019human}. Deep learning models can learn complicated spatial and temporal representations from data \citep{monir2021machine}. However, they frequently require large amounts of labelled data and significant computational resources \citep{tee2022close,gupta2021deep}. In recent years, there has been an increased interest in using generative models, particularly Restricted Boltzmann Machines (RBMs), for HAR. RBMs are a form of generative models that can be used to determine the underlying data distribution \citep{pienaar2019human,abdellaoui2020human}. This makes them ideal for applications like HAR, where the goal is to find the underlying patterns of human behaviours \citep{nie2015generative}. However, RBMs can be computationally expensive to train and may not be as capable of handling complex temporal data as deep learning models \citep{zhang2018overview}.

Given the diverse landscape of HAR models, each with its strengths and limitations, there is a clear need for a comprehensive comparison across these different model types. Such a comparison is essential to provide insights into how various models perform under different conditions, data complexities, and resource constraints. By understanding the advantages and limitations of traditional machine learning methods, deep learning architectures, and generative models, users can better assess which approaches are most suitable for different HAR scenarios and datasets. This evaluation is crucial for advancing the understanding of model selection in the context of HAR, particularly as the field continues to grow in importance for applications such as healthcare, sports science, and smart environments.

This study aims to address this critical need by conducting an extensive comparative analysis of different machine-learning models for human activity recognition. We investigate three different well-adopted model families, Traditional Machine Learning (TML) to non-traditional (Deep Learning) \citep{mauldin2018smartfall} and additionally Restricted Boltzmann Machines \citep{sedighi2020classification}. This paper contributes to the advancement of HAR research through the following key aspects:
\begin{itemize}

\item \textbf{Model Comparison Across Categories:}

This study conducts a detailed comparison of various machine learning approaches for HAR, including traditional models such as Decision Tree, Random Forest, Logistic Regression, Linear SVC, RBF-SVM, and KNN \citep{alhumayyani2021deep,shekhar2023human,mauldin2018smartfall}. Furthermore, it evaluates advanced deep learning techniques like CNN \citep{zeng2014convolutional,hammerla2016deep,mauldin2018smartfall,xu2021human,li2018comparison}, RNN \citep{hammerla2016deep,alhumayyani2021deep}, Long Short-Term Memory (LSTM) \citep{abdel2020st,alhumayyani2021deep,xia2020lstm,shekhar2023human,chen2016lstm,mekruksavanich2021lstm,pienaar2019human,hammerla2016deep}, Bi-directional LSTM (Bi-LSTM) \citep{hammerla2016deep,li2019bi,hernandez2019human,zhao2018deep}, and Gated Recurrent Units (GRU) \citep{alhumayyani2021deep,lu2022multichannel,dua2021multi}. It also incorporates ANN (\citep{alhumayyani2021deep,caya2019human,suto2018efficiency,gupta2022human}.
\item \textbf{Incorporation of generative architectures:}
To explore the power of generative models, we introduce the Restricted Boltzmann Machines (RBMs) as an additional model for comparison where we include Deep Belief Networks (DBNs) \citep{abdellaoui2020human}, Deep Boltzmann Machines (DBMs) \citep{sedighi2020classification,mocanu2015factored, gu2021survey}.
\item \textbf{Evaluation Across Key Datasets:}

We have gathered a collection of four diverse datasets for human activity recognition to evaluate the performance of the models in a variety of contexts. Each dataset presents a different activity scenario, environmental conditions, and data collection sources \citep{anguita2013public,de2018sensor}. By subjecting each model to these distinct datasets, we provide a comprehensive evaluation of their performance in the real world.

\item \textbf{Analysis of Model Strengths and Weaknesses:}

We present a detailed performance comparison utilizing criteria like accuracy, precision, recall, and F1-score. This allows us to identify the strengths and limits of classical, deep learning, and generative models \citep{hammerla2016deep,abdel2020st,zeng2014convolutional}.
\item \textbf{Real-World Application Guidance:}

The findings from our study offer valuable insights for practical deployment in fields such as healthcare, smart environments, and security \citep{mauldin2018smartfall,gupta2022human,alhumayyani2021deep}. By conducting a comprehensive analysis of different machine learning models on multiple datasets, our study bridges the gap between traditional machine learning, cutting-edge deep learning, and generative models. Our experiments provide valuable insights into the strengths and weaknesses of each model family and how they perform in different conditions.  Finally, our valuable findings will help researchers and developers in selecting appropriate models and building a reliable and efficient Human Activity Recognition system.
\end{itemize}

The rest of the paper is organized as follows. Section 2 presents the state of the art with a review of past HAR research, starting from traditional techniques of machine learning, the evolution of deep learning models and state-of-the-art results recently achieved with generative models. Section 3 introduces, in detail, the methodology of the research, data collection and preparation, and the selection and configuration of models. Section 4 details the performance of each model on several datasets. Section 5 describes  the limitations of the study and how this line of work can be advanced further. Finally, Section 6 summarizes the findings from our analysis and discusses their implications for real-world applications for HAR.  

\section{Related Works}
Several approaches have been developed to improve the accuracy and effectiveness of activity recognition systems \citep{ambati2021human,garcia2023new}. Human activity represents the different physical acts and gestures that individuals carry out in living their daily lives. These acts embody the spending of energy and can be anything from just walking, sitting, and eating to running, dancing, and even playing a musical instrument. However, it is human activity recognition that helps in identifying and categorizing this diverse human activity with the help of technology and various data analysis techniques. Among them, the most useful and popular medium for classifying human activities is Machine Learning models. These models learn and recognize patterns in data to make a difference between various activities based on unique movement and energy patterns. The early works related to HAR  focus on the classification and recognition of human activities using classical machine learning. \autoref{Rwork} provides an overview of prior studies, highlighting key models, datasets, and performance metrics that have shaped the development of HAR methodologies.

Recently, \cite{shekhar2023human} showed various machine learning models involving Decision Tree, Random Forest, Gradient Boosting DT, Logistic regression, Linear SVC, and RBF SVM classifier with the identification of activities such as sitting, walking, and standing in daily life. \cite{alhumayyani2021deep} proposed a model where they applied deep learning methodologies as feature extraction and traditional machine learning as a classifier for identifying human activities using smartphone sensors. In the paper \citep{ambati2021human,garcia2023new} experimented with three HAR datasets Pampap 2 (519,185 records), SWELL (189, 000 records), and MHealth (102,959 records) to apply the number of ML Techniques. Researchers like  \citep{garcia2023new} and \citep{randhawa2020human} presented the best accuracy model where SVM achieved 98.8\% and Random Forest outperformed from 74.39\% to 92.97\%. They also presented the training process of SVM and Random Forest, in which decision boundaries or construction of the ensemble models would effectively separate different activities. 
Traditional machine learning algorithms excel in scenarios where interpretability is essential, as they provide insights into the features and factors that contribute to classification decisions. So, the effectiveness of traditional machine learning lies in its ability to classify activity by learning patterns and relationships from labelled training data. However, traditional machine learning methods have some limitations to understanding complex patterns over time and automatically picking out details from data\citep{shakya2018comparative,lai2019comparison}. Deep learning has the potential to overcome these limitations by automatically learning from raw data and understanding complex patterns that change over time\citep{mohsen2021industry}. 
Recently, deep learning approaches have brought about a revolutionary change in the field of Human Activity recognition. 

\begin{table*}[p]
    \centering
    \normalsize
    \caption{Performance Comparison of Different AI Models for HAR}
    \resizebox{\textwidth}{!}{%
    \begin{tabular}{p{2.5cm}p{3cm}p{2.5cm}p{2.5cm}p{2.5cm}p{2.5cm}p{2.5cm}}
        \toprule
        Reference & AI Model & Dataset & \multicolumn{4}{c}{Performance}  \\
        & & & Accuracy & Precision & Recall & F1 Score  \\
        \midrule
        \citep{abdel2020st} & CNN, T-ResNet, T-DenNet, ResNet & UCI-HAR, WISDM & 96.16\%, 97.24\%, 96.38\%, 98.30\% & - & - & -   \\
        \midrule
        \citep{xia2020lstm} & CNN, DeepConvLSTM, LSTM-CNN & OPPORTUNITY, WISDM, UCI-HAR & - & - & - & 90.25\%, 92.32\%, 94.75\%  \\
        \midrule
        \citep{shekhar2023human} & Decision Tree, Random Forest, Gradient Boosting Decision Tree, Logistic Regression, Linear SVC, RBF SVM & UCI-HAR & 87\%, 91\%, 95\%, 92\%, 96.50\%, 97\% & - & - & -  \\
        \midrule
        \citep{chen2016lstm} & LSTM-based approach & WISDM & 92.1\%& - & - & -  \\
        \midrule
        \citep{hernandez2019human} & bidirectional long short-term memory (Bi-LSTM) & Human Activity Recognition Using Smartphones (UCI) & 92.67\% & - & - & -  \\
        \midrule
        \citep{lu2022multichannel} & Multichannel CNN-GRU & WISDM, UCI-HAR, PAMAP2 & 96.41\%, 96.67\%, 96.25\% & - & - & -  \\
        \midrule
        \citep{dua2021multi} & Multi-input CNN-GRU & UCI-HAR, WISDM, PAMAP2 & 96.20\%, 97.21\%, 95.27\% & - & - & -  \\
        \midrule
        \citep{abdellaoui2020human} & Deep Belief Networks (DBNs) & KTH and UIUC human action datasets & KTH = 94.83\%,  UIUC = 96\% & - & - & -  \\
        \midrule
        \citep{anguita2012human} & Multiclass SVM & UCI-HAR & 96\%& - & - & -  \\
        \midrule
        \citep{nie2015generative} & Local Interaction RBM (LRBM) & Extended Cohn-Kanade data set (CK+) & 88.6\% & - & - & 87.94\%  \\
        \midrule
        \citep{wan2020deep} & CNN, LSTM,
        BLSTM, MLP and SVM & UCI HAR and Pamap2 &CNN: 91.86\%
        LSTM: 87.94\%
        BLSTM: 89.46\%
        MLP:84.45\%
        SVM:87.29\%& CNN: 92.44\%
        LSTM: 87.83\%
        BLSTM: 89.80\%
        MLP:85.09\%
        SVM:87.61\% &CNN: 91.83\%
        LSTM: 86.83\%
        BLSTM: 89.19\%
        MLP:84.38\%
        SVM:87.05\% &  CNN: 92.05\%
        LSTM: 87.17\%
        BLSTM: 89.38\%
        MLP:84.54\%
        SVM:86.81\% \\
        \midrule
        \citep{mohsen2021industry} & STM, CNN, CNN-LSTM & WISDM & 
        LSTM:96.61\%
        CNN: 94.51\%
        CNN-LSTM: 97.76\% & LSTM: 96.57\%
        CNN: 94.83\%
        CNN-LSTM: 97.75\% & LSTM: 96.61\%
        CNN: 94.51\%
        CNN-LSTM: 97.77\% & LSTM: 96.57\%
        CNN: 94.61\%
        CNN-LSTM: 97.76\% \\
        \midrule
        \citep{mutegeki2020cnn} & CNN-LSTM & iSPL (Internal data set),  UCI 
        HAR & iSPL: 99.06\%
        UCI HAR: 92.13\% & - & - & - \\
        \midrule
        \citep{pang2021stacked} & Stacked Discriminant Feature Learning (SDFL) &  UCI-HAR &97\%  & 96.40\% & 96.30\%  \\
        \midrule
        \citep{zaki2020logistic} & K-Nearest Neighbour (KNN), 
        Naïve Bayes, Random Forest, Gradient Boosting and Logistic Regression & UCI-HAR, HAPT & UCI-HAR:
        KNN:89.1
        Gaussian NB:77.0\%
        RF:92.4\%
        GB:93.7\%
        LR:96.1\%
        HAPT:
        KNN:87.2\%
        Gaussian NB:74.7\%
        RF:90.8\%
        GB:91.7\%
        LR:94.5\% & - & - & -  \\
        \midrule
        \citep{mohsen2021human} & KNN& UCI-HAR & 90.46\% & 90.96\% & 90.46\% & 90.37\% \\
        \midrule
        \citep{fang2024casnn} & continuous adaptive spiking neural 
network (CASNN) & UCI-HAR, UniMiB SHAR, HHAR  & UCI-HAR:  94.49\%
UniMiB SHAR: 97.37\%
HHAR: 89.70\% & - & - & -  \\
        \midrule
        \citep{geravesh2023artificial} & GRU & WISDM & 97.08\%& 97.11\%, & 97.09\% & 97.10\%  \\
    \bottomrule
    \end{tabular}}
    \label{Rwork}
\end{table*}

 \cite{alhumayyani2021deep} and  \cite{wan2020deep} utilized a wide range of DL models including CNN, RNN including LSTM, Bi-LSTM and GRU and Multi-layer Perception (MLP) which is a forward structured ANN. \cite{zeng2014convolutional} developed a CNN-based method to capture local dependence and preserve feature scale invariant to recognize human activities and the proposed model outperformed the state-of-the-art methods. Another exploration by \cite{pienaar2019human} analyzed RNN with LSTM to design LSTM architecture where six main activities were considered: sitting, Jogging, Standing, Walking, upstairs and downstairs. \citep{lu2022multichannel} used tree benchmark dataset WISDM (Wireless Sensor Data Mining) (Single Sensor, 36 participants and 6 activities), UCI-HAR (Multi-sensor, 30 Volunteers, 6 Activities) and PAMAP2 (Multi-sensor, 9 subject and 12 activities) to develop multi-channel CNN-GRU model to classify Human Activity.  \citep{chen2016lstm}  and  \citep{mutegeki2020cnn} achieved promising results applying deep learning approaches where LSTM-based approaches achieved 92\% accuracy and CNN-LSTM achieved 99\% accuracy on an internal dataset and achieved 99\% accuracy on the UCI HAR dataset. Deep Learning can automatically understand complex things, extract features from raw data, and recognise hidden patterns. It can also handle sequential dependencies, improve recognition performance, learn hierarchical representations, and reduce manual feature engineering. Although deep learning is good at identifying hidden patterns, it can face difficulties when dealing with tasks that involve complex temporal dependencies and high dimensional data\citep{ronao2016human}. This is where Deep Boltzmann Networks were added in our experiments. DBNs can model complex relationships within data and handle multidimensional inputs.
 
Deep models employed in HAR include RBM-based and generative models. These models consist of multiple layers of RBMs stacked on top of each other\citep{fang2014recognizing}. According to \citep{gu2021survey} survey paper, RBMs serve as the foundation for deep learning models such as DBNs, DBMs, and Convolutional Boltzmann Machines (CBMs). \citep{abdellaoui2020human} proposed a DBNs model that can extract features from videos and classify human activity. \citep{sedighi2020classification}  conducted an experiment using the SBHAR dataset to recognize human activity; their results showed that DBNs achieved 98.25\% accuracy in training data and 93.01\% accuracy in testing data. RBMs help to enrich our understanding of HAR by revealing previously unseen patterns and relationships. Therefore, these generative models in HAR research encourage researchers to reassess their viewpoints and adopt a renewed perspective that aligns with these models.

\section{Research Design and Data Collection}
This section outlines the research strategy and data collection methods used to evaluate various machine learning models for HAR. We used five benchmark datasets: UCI-HAR, OPPORTUNITY, PAMAP2, WISDM, and Berkeley MHAD, each chosen for its distinct features and problems. The models examined include classical machine learning techniques like Decision Trees and Random Forests, as well as deep learning architectures like CNN and RBMs, specifically DBNs and DBMs. We discuss the performance indicators used in our analysis, including accuracy, precision, recall, and F1-score, and provide a methodology for evaluating the models' capabilities in real-world HAR applications.

\subsection{Datasets}

One of the critical parts of our research is to select appropriate datasets. In the data collection process for our comprehensive comparative study, we focused on the dataset's complexity and whether the datasets are already extensively solved or not for classification solutions. To ensure meaningful comparison, we chose four publicly available datasets to conduct our experiments. The following sections provide a clear picture of the datasets to describe the basics of the four datasets. 

\subsubsection{UCI-HAR} \citep{anguita2012human}
The UCI-HAR is a data collection from 30 users between 19 to 48 years. Each person performed six activities (walking, walking upstairs, walking downstairs, sitting, standing, laying) with 561 features belonging to accelerometer and gyroscope sensors of a smartphone (Samsung Galaxy S11) wearing on the waist \citep{chaurasia2019ai,pang2021stacked}. The smartphone’s built-in accelerometer and gyroscope recorded data and the dataset measurement from triaxial accelerometers and gyroscope sensors. The data was recorded at a frequency of 50 Hz. The obtained dataset was divided into 2 sets where training data was generated by 70\% of the users and 30\% 
test data \citep{kann2023evaluation}. According to the statistical analysis, the dataset contains a large amount of data, totalling 748,406 individual samples. \autoref{T1} contains classes of UCI-HAR and their proportions of data for each class
\begin{table}[htbp]
  \centering
  \begin{tabular}{|l|c|}
    \hline
    Class               & Proportion (\%) \\
    \hline
    Laying              & 18.88 \\
    \hline
    Standing            & 18.51 \\
    \hline
    Sitting             & 17.25 \\
    \hline
    Walking             & 16.72 \\
    \hline
    Walking upstairs    & 14.99 \\
    \hline
    Walking downstairs  & 13.65 \\
    \hline
  \end{tabular}
  \caption{ Proportions of data for each class}
  \label{T1}
\end{table}

\subsubsection{Opportunity} \citep{roggen2012opportunity}

The opportunity dataset is a multimodal dataset designed for Human Activity Recognition from wearable, mobile and ambient sensors to benchmark human activity recognition algorithms (classification, automatic data segmentation, sensor fusion, feature extraction etc.)\citep{sagha2011benchmarking}. The sensor data was collected from subjects wearing a set of wearable devices including wrist, chest, hip, and dominant forearm. The data was collected at a frequency of 30 HZ. The dataset includes precisely annotated data from a group of 4 subjects both male and female to support the perception and learning of various human activities such as short action, gesture, modes of locomotion, and high-level behavior\citep{gioanni2016opportunistic}. Overall, the opportunity dataset contains data from 35 activities categorized into 13 low-level. The dataset includes data collected from 23 body-worn sensors, 12 object sensors and 21 ambient sensors\citep{foudeh2016testing}. 

\subsubsection{PAMAP 2} \citep{misc_pamap2_physical_activity_monitoring_231} 

The PAMAP 2 dataset includes measurements from different sensors such as tri-axial accelerometer, gyroscope, Magnetometer, and heart rate data. This dataset is a collection of data collected from 9 subjects performing 18 different physical activities including walking, running, and climbing stairs. The sensor data was collected from subjects while wearing 3 IMUs devices on the arm, chest, and ankle and the heart rate data was collected from an HR-monitor\citep{tokmak2022unveiling}. The data was recorded at a frequency of 100 Hz. The PAMAP 2 \autoref{T2} dataset is a valuable resource for researchers working on activity recognition. The dataset can be developed algorithms for data processing, segmentation, feature extraction and classification\citep{reiss2012creating}.  
\begin{table}[htbp]
  \centering
  \begin{tabular}{|l|c|}
    \hline
    Class               & Proportion (\%) \\
    \hline
    Walking             & 12.29\% \\
     \hline
    Ironing             & 12.29\% \\
     \hline
    Lying               & 9.90\% \\
     \hline
    Standing            & 9.78\% \\
     \hline
    Nordic walking      & 9.68\% \\
     \hline
    Sitting             & 9.53\% \\
     \hline
    Vacuum-cleaning     & 9.02\% \\
     \hline
    Cycling             & 8.47\% \\
     \hline
    Ascending stairs    & 6.03\% \\
     \hline
    Descending stairs   & 5.40\% \\
     \hline
    Running             & 5.05\% \\
     \hline
    Rope jumping        & 2.54\% \\
    
    \hline
  \end{tabular}
  \caption{ Proportions of data for each class}
  \label{T2}
\end{table}

\subsubsection{WISDM (Wireless Sensor Data Mining)} \citep{kwapisz2010activity}

WISDM (Wireless Sensor Data Mining) acquired from WISDM Lab is a project of Fordham University that is focused on collecting and mining data from accelerometers and gyroscopes of phones and watches. The dataset contains data collected from 36 subjects, each of whom was asked to perform six types of human activities including upstairs, downstairs walking, jogging, sitting, and standing for specific periods. Accelerometer data measured through different dimensions X, Y and Z axes. The data was recorded at a frequency of 20 Hz. This WISDM dataset, see \autoref{T3}, is used by researchers around the world to develop and evaluate algorithms for activity recognition, fall detection and other applications of sensor data mining. This dataset is well organized and relatively large which allows for robust training of machine learning models \citep{oluwalade2021human}. 

\begin{table}[htbp]
  \centering
  \begin{tabular}{|l|c|}
    \hline
    Class               & Proportion(\%) \\
    \hline
    Downstairs          & 38.93\% \\
    \hline
    Jogging             & 30.23\% \\
    \hline
    Sitting             & 11.42\% \\
    \hline
    Standing            & 9.33\% \\
    \hline
    Upstairs            & 5.58\% \\
    \hline
    Walking             & 4.51\% \\
    \hline
  \end{tabular}
  \caption {Proportions of data for each class}
  \label{T3}
\end{table}

\noindent Berkeley MHAD (Multimodal Human Action Database)\citep{golestani2020berkeley}  

\subsubsection{The Berkeley MHAD (Multimodal Human Action Database)} \autoref{T4} is a database of human activity data collected using multiple devices including RGB Cameras, a depth sensor (Kinect V1), an inertial sensor (3-axis accelerometer) a thermal sensor and a microphone. The dataset contains 12 actions including multiple sensor modalities, depth images, infrared images, skeleton joint positions, and inertial sensor data performed by 7 male and 5 female subjects in the range of 23-30 years of age. 
\begin{table}[htbp]
  \centering
  \begin{tabular}{|l|c|}
    \hline
    Class                   & Proportion(\%)\\
    \hline
    Jumping in place        & 5.93\% \\
     \hline
    Jumping jacks           & 7.77\% \\
     \hline
    Bending                 & 18.53\% \\
     \hline
    Punching                & 9.91\% \\
     \hline
    Waving (two hands)      & 10.12\% \\
     \hline
    Waving (one hand)       & 10.58\% \\
     \hline
    Clapping Hands          & 5.24\% \\
     \hline
    Throwing a ball         & 3.58\% \\
     \hline
    Sit Down then stand up  & 20.74\% \\
     \hline
    Sit down                & 3.58\% \\
     \hline
    Stand up                & 2.88\% \\
     \hline
    T-pose                  & 1.15\% \\
    \hline
  \end{tabular}
  \caption{ Proportions of different activities.}
  \label{T4}
\end{table}
Each subject performed each action 5 times, yielding about 660 action sequences which correspond to about 82 minutes of total recording time at a frequency of 30 Hz. The data was collected both indoors and outdoors \citep{ofli2013berkeley,ng2017action}. This dataset is a comprehensive dataset designed to support researchers in various areas including human activity recognition, computer vision, and machine learning.

\subsection{Models}
\subsubsection{Decision Tree} 

\begin{figure}[!ht]
  \centering
  \includegraphics[width=0.5\textwidth]{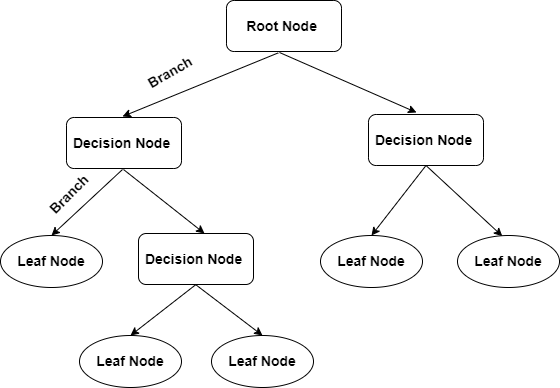}
  \caption{A general illustration of a decision tree.}
  \label{F1}
\end{figure}

Decision Trees are a supervised learning algorithm used for both classification and regression tasks, effective in classifying activities like walking, running, or sitting by creating simple rules based on sensor data features like acceleration and orientation\citep{maswadi2021human}. Decision trees \autoref{F1} are built in a top-down manner where each node contains a single value, and the root node represents the entire dataset. Each branch represents the outcome of the decision. Decision trees work by splitting data into subsets recursively based on chosen criteria depending on the trimming process. Splitting data can be done by impurity measurement such as GINI index and entropy\citep{nurwulan2021human,sanchez2018decision}. 

\noindent The Decision tree algorithm starts the root node and follows the branches down the tree until it reaches a leaf node to predict the values for new data\citep{fan2013human}. Let's express the concept of a decision tree through the following equation: 

\begin{equation}
y = f(x_1, x_2, \ldots, x_n) \label{eq:decision_tree}
\end{equation}
where
\begin{table}[!ht]
\centering
\begin{tabular}{p{0.9\columnwidth}}
$y$ is the target variable (class or regression value) \\
$f$ is the decision tree function \\
$x_1, x_2, \ldots, x_n$ are the input features \\
\end{tabular}
\end{table}
 
\subsubsection{Random Forest}  
\begin{figure}[H]
  \centering
  \includegraphics[width=0.3\textwidth]{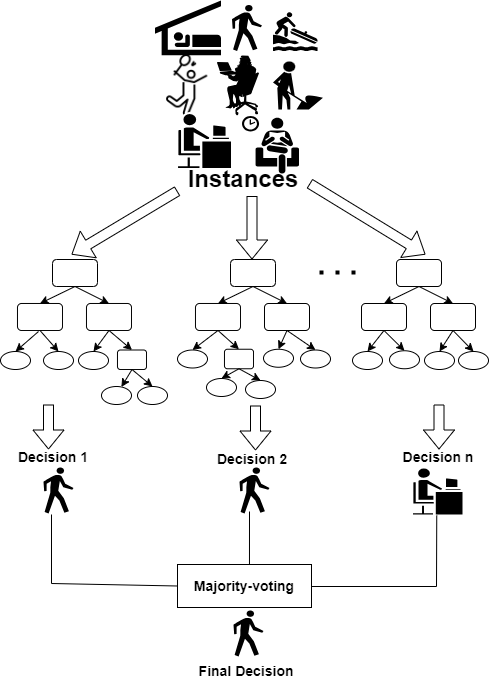}
  \caption{A visual illustration of the Random Forest}
  \label{F2}
\end{figure}

Breiman \citep{breiman2001random} developed Random Forest which is an ensemble learning method that improves robustness and accuracy. It is particularly useful for HAR due to its ability to manage complex sensor data and reduce overfitting by combining multiple decision trees as described \autoref{F2}. The random forest algorithm involves a multi-step process \citep{maswadi2021human}.
\paragraph{Sampling}
This is the key step where subsets are selected from datasets. Especially, one dataset contains K number of records, n random records chosen with a subset of m features from k records. This process introduces randomness and diversity into the model.

\noindent Construction  

\noindent After sampling, individual decisions that constructed from each subset which is built with n random records and m features.\newline 
Output  

\noindent Every decision tree generates its own set of predictions or outputs based on the data it was trained on.\newline  
Voting  

\noindent For the final prediction majority voting applies for the classification and averaged technique used for regression. 
\begin{equation}
y_{\hat{k}} = \text{argmax}_k \text{avg}(f_k(x)),\ k = 1, \ldots, K
\end{equation}

\begin{table}[!ht]
\centering
\begin{tabular}{p{0.9\columnwidth}}
$y_{\hat{k}}$ is the predicted target variable (class or regression value) \\
$(f_k(x))$ is the prediction of the kth decision tree in the random forest \\
$K$ is the number of decision trees in the random forest 
\end{tabular}
\end{table}

\subsubsection{Logistic regression }

Logistic regression is a predictive analysis like other regression analysis \citep{zaki2020logistic,madarshahian2015human}. It describes data and explains the connection between one dependent binary variable meaning that it can take on only two variables such as “1” or “0”, “yes” or “no”, “true” or “false”, “sick” or “healthy”, and in the context of HAR, it distinguishes between two activities like sitting versus standing or walking versus running. Logistic regression starts by fitting a linear regression model to the data. The linear regression model predicts a continuous value, which is passed through a logistic function or sigmoid function S-shaped curve that takes a real number as input and outputs a value between 0 and 1 to produce a probability\citep{cramer2002origins}. 
Equation (3) represents a mathematical expression used in a machine learning context, specifically for logistic regression with L2 regularization. It's trying to find the best set of weights $W^*$ to make predictions based on data.
\begin{equation}
W^*=\arg _w \min \left[\sum_{i=1}^n \log \left(1+\exp \left(-y_i w^T x_i\right)\right)+\lambda w^T w\right]
\end{equation}

\subsubsection{Linear SVC }

Linear SVC (Support Vector Classification) is a supervised classification algorithm extending Support Vector Machine (SVM) capabilities for scenarios where data can be neatly divided to be used for binary and multiclass classification tasks\citep{ladicky2011locally}. It is particularly effective for classifying human activities based on sensor data, such as accelerometer, gyroscope, and magnetometer readings and ensuring a robust classification performance across various tasks.
It works by finding a hyperplane that maximizes the distance between two classes, effectively dividing the data so that each side contains points from only one class \citep{tang2013deep}. Once the hyperplane is determined, specific features of a new instance can be input into the classification model to predict its class.
For linear kernels, linear SVC is a faster implementation of SVM \citep{apostolidis2015svm}. The equation is finding the best weights $(w*)$ and bias $(b*)$ for an SVM model. It tries to strike a balance between having small weights (to prevent overfitting) and minimizing the classification error (to ensure good performance in classifying data points), where the trade-off is controlled by the regularization parameter $c$.
\begin{equation}
\left(w^*, b^*\right)=\operatorname{argmin}_{w, b} \frac{\|w\|}{2}+c \cdot \frac{1}{n} \sum_{i=1}^n \xi_i
\end{equation}

\subsubsection{RBF SVM   } 

Radial Basis Function Support Vector Machine (RBF SVM) is a type of SVM algorithm that uses a radial basis function as a kernel function for binary classification and it is recommended in scenarios where data is linearly inseparable or non-linear. Using the mapping technique input data transform into a higher dimensional space where the data become linearly separable\citep{ijraset2020human}. After mapping using the RBF kernel, a SVM classification is used to find a hyperplane and then perform the classification using the basic Idea of Linear SVC. RBF SVM performs well in non-linear and bidimensional scenarios\citep{thurnhofer2020radial,patle2013svm}. RBF SVM is particularly effective due to its ability to capture complex relationships in sensor data from smartphones and wearables. By maximizing the non-linear transformation capabilities of the RBF kernel, HAR systems can accurately classify various human activities, such as walking, running, and sitting, even in diverse environments and conditions. Equation 5 plays an important role in computing hyperplanes.  
\begin{equation}
\max _{\alpha_{\mathrm{i}}} \sum_{\mathrm{i}=1}^{\mathrm{n}} \sum_{\mathrm{j}=1}^{\mathrm{n}} \alpha_i \cdot \alpha_{\mathrm{j}} \cdot \mathrm{y}_{\mathrm{i}} \cdot \mathrm{y}_{\mathrm{j}} \cdot \operatorname{rbf} \_\operatorname{kernel}\left(\mathrm{x}_{\mathrm{i}}, \mathrm{x}_{\mathrm{j}}\right)
\end{equation}
where $\alpha_i$ and $\alpha_j$ are the Lagrange multipliers, $y_i$ and $y_j$ are the labels of points \(x_i\) and \(x_j\), and  \(\text{rbf\_kernel}(x_i, x_j)\) is the RBF kernel.

\subsubsection{K-nearest Neighbour (KNN)  } 

\begin{figure}[H]
  \centering
  \includegraphics[width=0.5\textwidth]{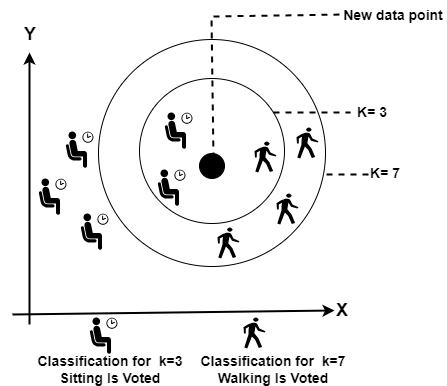}
  \caption{A visual illustration of KNN algorithm}
  \label{F3}
\end{figure}

KNN is a versatile supervised machine learning algorithm that can be used for both classification and regression tasks, including applications in HAR. It works by finding similar data points from the same labels or value and nearest data from different labels, see \autoref{F3}.
K-nearest Neighbour tries to make accurate predictions by finding the k most similar instance in the training set to a new instance and predicts the label of the new instance\citep{mohsen2021human,wang2021application,kaghyan2012activity}. When classifying a new activity, KNN examines nearby data points of various labels (e.g., walking, running, sitting) and assigns the most common label among these neighbours to the new instance. However, it measures the distance between the test data point and every point within the training dataset. 

\subsubsection{Convolution Neural Network (CNN)}
\begin{figure}[H]
  \centering
  \includegraphics[width=0.5\textwidth]{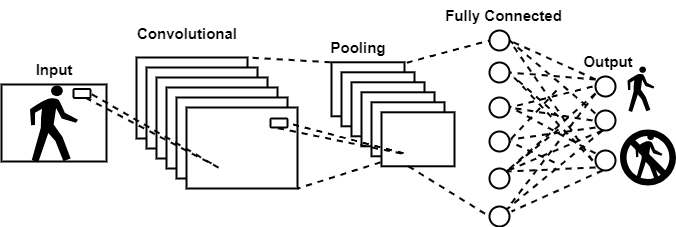}
  \caption{Visual representation of different layers in a CNN.}
  \label{F4}
\end{figure}

CNN is an artificial neural network specially designed to learn directly from data\citep{bevilacqua2019human,albawi2017understanding}. It is particularly useful for automatically extracting spatial features from raw sensor data and capturing patterns in time-series signals. CNN architecture is structured with various specialized layers, each serving a distinct role in the network’s architecture. The layers include an input layer, an output layer, convolutional layers, pooling layers, and fully connected layers \citep{o2015introductionshi} \autoref{F4}.  
\noindent Convolutional layers \newline 
The convolutional layers extract complex features from the input data by applying a convolution operation to the input data which is a mathematical operation. 

\noindent Pooling Layers  \newline
These layers are used to reduce the size of the features produced by the convolutional layers. By this action, the number of parameters is reduced in the network to make the network more computationally efficient.  

\noindent Fully connected layers \newline 
Just before the output layer, the final fully connected layers combine the outcomes of convolutional and pooling layers and make a prediction. 

\noindent Output layer \newline
The output layer generates classification results on the outputs of the fully connected layers \newline

\subsubsection{Recurrent Neural Network (RNN)}
  
\begin{figure}[H]
  \centering
  \includegraphics[width=0.3\textwidth]{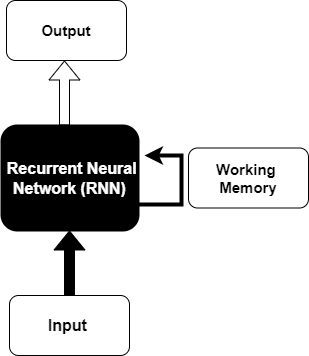}
  \caption{RNNs utilize their internal state, or memory, to analyze sequences of inputs.}
  \label{F5}
\end{figure}

RNN \autoref{F5} is designed for time series or sequential data, where data points have a temporal order. In standard neural networks, all inputs and outputs are considered independent of one another \citep{singh2017human}. 
However, in scenarios like Human Activity Recognition (HAR), where sensor data is captured in continuous sequences, such as accelerometer and gyroscope readings, past information is crucial for predicting future activities \citep{paydarfar2020human}. The hidden state of the RNN plays a key role in maintaining context from prior inputs, enabling the network to recognize patterns in sequential data, such as distinguishing between walking, running, or sitting, based on the flow of sensor signals \citep{sherstinsky2020fundamentals}. RNNs are an effective solution for processing and understanding temporal relationships in HAR datasets. There are three main components of RNN:

\noindent Input layer  \newline
This layer receives the raw data or features. Its purpose is to pass this data to the subsequent layers for processing. 

\noindent Hidden layers \newline
These layers perform the core computation of the neural network. They are responsible for processing the input data and learning the long-term dependencies in the data. 

\noindent Output layer  \newline
This layer is the final layer in the neural network. This layer is responsible for producing the network's prediction and classification. 

\subsubsection{Long Short-term Memory (LSTM) }

LSTM networks, designed by Hochreiter and Schmidhuber \citep{hochreiter1997long}, were introduced to address the shortcomings of standard RNNs in handling long-term dependencies in sequential data. While RNNs are capable of capturing recent information, they often struggle to retain and utilize data stored in long-term memory, leading to diminished performance in tasks requiring an understanding of extended sequences. LSTM overcomes this limitation by incorporating feedback connections, allowing it to process entire sequences of data, not just individual points \citep{li2022human,siami2019performance}. This makes LSTM particularly efficient for tasks involving sequential data, such as text, speech, time series, and Human Activity Recognition (HAR). In HAR, where sensor data from devices like smartphones and wearables is gathered over time, LSTM's ability to retain and process long-term dependencies helps in accurately identifying activities over extended periods. Memory cells at the LSTM architecture are core to enable the network to retain crucial information over long sequences, making it a powerful tool for sequential data prediction and understanding. These memory cells have three key components: 
\noindent Input Gate    \newline
The input gate regulates how much of the new input is allowed to enter the cell.  

\noindent Hidden layers \newline
These layers perform the core computation of the neural network. They are responsible for processing the input data and learning the long-term dependencies in the data. 

\noindent Output layer  \newline
This layer is the final layer in the neural network. This layer is responsible for producing the network's prediction and classification. 
\begin{figure}[H]
  \centering
  \includegraphics[width=0.4\textwidth]{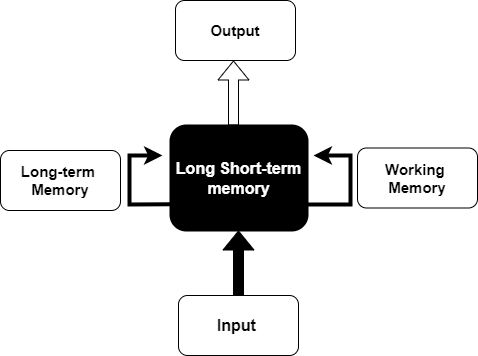}
  \caption{LSTM networks represent a specialized variant of RNNs}
  \label{F6}
\end{figure}

These three gates are implemented using sigmoid functions. This sigmoid function is a non-linear function that outputs between 0 and 1. The gate is closed when the value is 0 and a value of 1 means it is open

\subsubsection{BiLSTM (Bi-directional Long Short-Term Memory)  }

\begin{figure}[H]
  \centering
  \includegraphics[width=0.5\textwidth]{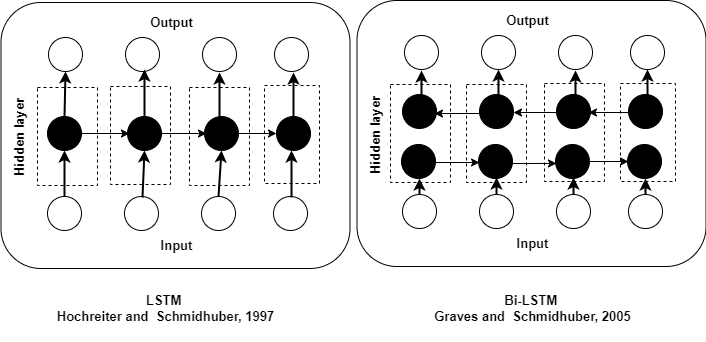}
  \caption{The architectures of LSTM and Bi-LSTM}
  \label{F7}
\end{figure}

BiLSTM is a type of RNN and an extension of the LSTM architecture \autoref{F7} designed to handle sequential data. Unlike traditional LSTM, which processes sequences in a single direction either forward or backward, BiLSTM processes the sequence in both directions \citep{li2019bi}. It uses two LSTM layers: one processes the sequence from past to future (forward), and the other processes it from future to past (backwards). This bidirectional approach allows BiLSTM to capture dependencies and context from both previous and future steps in the sequence, making it particularly effective in tasks where understanding long-term relationships is essential. In particular, in fields like Human Activity Recognition (HAR), where sensor data from wearable devices is analyzed over time, BiLSTM can extract valuable information from both earlier and later time steps, enabling more accurate predictions of activities such as walking or running. The architecture’s key feature is the fusion of information from both directions, with the outputs of the forward and backward LSTM layers integrated to produce the final result \citep{li2022human,siami2019performance}. This makes BiLSTM well suited for  HAR tasks, as well as other domains like natural language processing and speech recognition.
\subsubsection{Gated recurrent units (GRUs)   }

GRUs are a type of RNN and a simplified version of LSTM \autoref{F8} introduced by Jun-Young Chung et al. \citep{chung2014empirical}. Both GRUs and LSTMs are widely used in time-series and sequence-based tasks, including HAR, where modelling temporal dependencies in data is critical. GRUs, in particular, are effective in HAR because they can efficiently process sequential data. The gating mechanisms in GRUs allow the model to control the flow of information through the network. These gates decide what information should be kept, forgotten, or updated, which is vital in HAR, where long-term dependencies need to be captured to correctly recognize activities. GRUs achieve this with fewer parameters than LSTM, making them computationally efficient while maintaining competitive performance \citep{mohsen2023recognition}. GRUs use two gating Mechanisms:
\begin{figure}[H]
  \centering
  \includegraphics[width=0.4\textwidth]{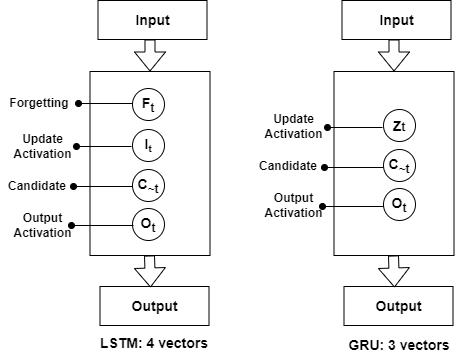}
  \caption{A visual comparison of GRU and LSTM}
  \label{F8}
\end{figure}
\noindent Reset Gate: \newline
This gate determines how much of the previous state should be kept or forgotten. 

\noindent Update Gate  \newline
This update gate controls which parts of the current state should be updated. 

These gates allow the GRU to effect model sequential data by retaining relevant information from the past and removing irrelevant information.\newline 
\subsubsection{Artificial Neural Network (ANN)  }

ANNs are machine learning models inspired by the human brain, consisting of interconnected nodes or artificial neurons that process information like biological neurons do \citep{grossi2007introduction, maind2014research}. In HAR, ANNs analyze data from sensors in smartphones or wearables to identify activities such as walking or running. Information flows through various layers in an ANN, with each connection having an adjustable weight. During the learning process, ANNs optimize these weights to minimize the difference between predicted and actual outputs, improving classification accuracy in HAR tasks \citep{geravesh2023artificial}. ANNs are typically structured into three main layers \citep{kwon2018recognition} \autoref{F9}shown below.   
\begin{equation}
M_k=f\left(\left[\sum_{j-1}^{l_k} w_{k j} f\left(\sum_{i=1}^u w_{j i} x_i+w_{j 0}\right]+w_{k 0}\right)\right.
\end{equation}
The activation of the neuron, denoted by $M_k$, is a function of the weighted sum of its inputs, $x_i$, where $k$ is the index of the input and $j$ is the index of the neuron. The weights between the neuron $j$ and the neuron $i$ and between the neuron $j$ and the output $k$ are denoted by $W_{j_i}$ and $W_{k_j}$, respectively. 
\begin{figure}[H]
  \centering
  \includegraphics[width=0.4\textwidth]{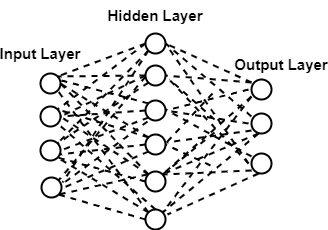}
  \caption{Architecture of ANN layers}
  \label{F9}
\end{figure}
\noindent Input Layer: \newline
Input layers receive the input data, such as an image or a text string process it and deliver it to the next layer. \newline
\noindent Hidden Layers:
There are one or more hidden layers in ANN architecture to perform complex computations on the input data and learn to extract features that are useful for the task. \newline 
\noindent Output Layer \newline 
The output layer produces the final prediction on classification. 
ANN is trained by feeding them large amounts of data and allowing them to the relationships between the inputs and designed output. After training, it can be used to make predictions or decisions on new data.\newline
\subsubsection{Deep Belief Networks (DBNs)  }
\begin{figure}[H]
  \centering
  \includegraphics[width=0.3\textwidth]{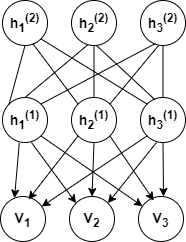}
  \caption{The structure of DBNs model}
  \label{F10}
\end{figure}

DBNs are a generative graphical model or a class of deep neural networks with many hidden layers consisting of visible (input) units, hidden units, and output units\citep{hinton2006fast,sedighi2020classification}. In DBNs, the visible units (input) can take binary or real values whereas the hidden units are often binary. Generally, units in one layer are connected to the adjacent layers, this term except in a sparse DBN. The connection between the top two layers is not directed, all the other layers are directed\citep{gu2021survey} \autoref{F10} Constructing a DBN involves a process of layering Multiple Restricted Boltzmann Machine (RBMs) a top of another layer. This stacking of RBMS forms the foundation of DBN architecture. RBMs are a type of generative model that can learn to represent the probability distribution of a set of data\citep{bondarenko2013research}.
\noindent Therefore, DBN is particularly powerful in modelling complex patterns and extracting meaningful features from large and high-dimensional datasets. DBN has demonstrated success in achieving state-of-the-art results in various domains including Human Activity Recognition, natural language processing, and speech and image recognition. \newline
\subsubsection{Deep Boltzmann Machines (DBMs)}

\begin{figure}[H]
  \centering
  \includegraphics[width=0.3\textwidth]{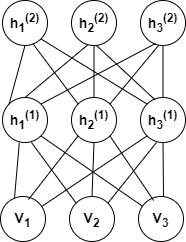}
  \caption{The structure of DBMs model}
  \label{F11}
\end{figure}

DBMs are generative, unsupervised deep learning models comprising three layers that learn complex representations and high-level features, making them ideal for HAR. They can capture intricate patterns in sequential data from sensors, such as smartphones and wearables \citep{feng2023monotone}. Similar to DBNs, DBMs consist of multiple layers of RBMs. However, a key distinction is that in DBMs, all connections between the RBMs are undirected \citep{gu2021survey}  \autoref{F11}, allowing for more flexible learning of feature hierarchies. 
\noindent This undirected nature enables DBMs to better represent the complex dependencies \citep{popa2018complex,goodfellow2013multi} between different activities and their corresponding sensor readings in HAR tasks. By leveraging their ability to learn high-level features from raw sensor data, DBMs can improve the accuracy of activity classification and enhance the interpretability of the underlying patterns, ultimately facilitating more robust HAR systems.

\subsection{Evaluation measures for performance}
The evaluation of a Human Activity Recognition Performance is crucial in measuring its effectiveness. In this experimental study, evaluating all the models for human activity recognition, we considered some commonly used evaluation matrices such as accuracy, recall, precision and F1-score. These selected methods offer a significant perspective allowing us to gain a comprehensive understanding of the models’ performance.  

\subsubsection{Accuracy}
 Accuracy is a fundamental performance metric that describes how the model performs accuracy on a given test data to provide the number of correctly classified samples over the number of samples. The model's performance depends on a high accuracy score which ranges from 0 to 100. 
\begin{equation}
\text { Accuracy }=\frac{\mathrm{TP}+\mathrm{TN}}{\mathrm{TP}+\mathrm{TN}+\mathrm{FP}+\mathrm{FN}}.
\end{equation}
\subsubsection{Precision } 
Precision is defined as the percentage of correctly classified or predicted samples over the total number of classified or predicted positive (True and False) samples. In terms of Human Activity Recognition (HAR), precision helps identify specific activity, reducing false positive classification. 
\begin{equation}
\text { Precision }=\frac{TP}{TP+FP}
\end{equation}
\subsubsection{Recall   } 
Recall or sensitivity is the percentage of correctly predicted positive samples to the total actual samples. A higher recall indicates that the model used in our experiment is better at capturing all instances of the positive class in HAR. 
\begin{equation}
\text { Recall }=\frac{TP}{TP+FN}
\end{equation}
\subsubsection{F1-score    } 
 F1-score combines precision and recall developing a balanced metric that considers both false positive and false negative. 
\begin{equation}
\text { F1 Score }= \frac{2 * \text { Precision } * \text { Recall }}{\text { Precision }+ \text { Recall }}
\end{equation}
\noindent F1-score calculated as the weighted average of precision and recall each given a weight of 2. 

\subsubsection{Confusion Matrix   } 
 Confusion matrix is a table matrix that visualizes the full performance of the classification models by comparing its predicted labels with the actual labels of the data. In the table, columns represent the predicted classes, and rows represent the actual classes. 

\section{Performance Metrics and Results Analysis}
In this section, we provide the findings from our comparative analysis of the machine learning models used for HAR across the selected benchmark datasets. We thoroughly analyse the performance results, focusing on key metrics such as accuracy, precision, recall, and F1-score, which are critical for determining model success in real-world scenarios. The topic focuses on the differences in performance between classical machine learning models, deep learning architectures, and RBMs, emphasizing their strengths and limitations concerning dataset features. By examining these findings, we hope to better understand the implications for selecting optimal models for HAR tasks and suggest possible areas for future research and advancement in this quickly expanding field.

\subsection{Result for UCI-HAR dataset}

\noindent UCI-HAR dataset includes 10,299 samples. From this number, we partitioned 7,352 samples for the training set and reserved 2,947 samples for the testing set. After sample partitioning, we trained classical machine learning models, deep learning models, and Restricted Boltzmann Machines. Subsequently, we evaluated the performance of each model using standard performance metrics which are explained in Subsection 3.3.  

Several classical machine learning including RBF SVM, Linear SVC and Logistic Regression achieved the highest performance exceeding scores of 0.95 across all metrics: accuracy, precision, recall and F1 score. Random forest also performed well with scores surpassing 0.92, as shown in \autoref{T5}. However,  KNN and decision tree were slightly less accurate with scores of over 0.85.  

\begin{table}[htbp]
  \centering
  \small 
 
  \setlength{\tabcolsep}{4pt} 
  \begin{tabular}{p{0.19\columnwidth}cccc} 
    \toprule
    \textbf{Model Name} & \textbf{Precision} & \textbf{Recall} & \textbf{F1-Score} & \textbf{Accuracy} \\
    \midrule
    Decision Tree       & 0.86               & 0.85             & 0.85               & 0.86              \\
    Random Forest       & 0.92               & 0.92             & 0.92               & 0.92              \\
    Logistic Regression & 0.95               & 0.95             & 0.95               & 0.95              \\
    Linear SVC          & 0.96               & 0.96             & 0.96               & 0.96              \\
    RBF SVM             & 0.95               & 0.94             & 0.95               & 0.95              \\
    K-nearest Neighbor  & 0.90               & 0.89             & 0.89               & 0.90              \\
    \bottomrule
  \end{tabular}
   \caption{Performance Metrics of Classical Models}
   \label{T5}
\end{table}

Among deep learning models, CNN and ANN performed as the top performers achieving exceptional scores above 0.94. However, RNN, LSTM, Bi-directional LSTM, and GRUs demonstrated lower performances with scores ranging from 0.78 to 0.85, see \autoref{T6}. 

\begin{table}[htbp]
  \centering
  \small 
  \setlength{\tabcolsep}{4pt} 
  \begin{tabular}{p{0.16\columnwidth}cccc} 
    \toprule
    \textbf{Model Name} & \textbf{Precision} & \textbf{Recall} & \textbf{F1-Score} & \textbf{Accuracy} \\
    \midrule
    CNN         & 0.95 & 0.95 & 0.95 & 0.95 \\
    RNN             & 0.82 & 0.78 & 0.78 & 0.79 \\
    LSTM              & 0.83 & 0.82 & 0.82 & 0.82 \\
    Bi-LSTM & 0.85 & 0.84 & 0.84 & 0.85 \\
    GRU                & 0.83 & 0.82 & 0.82 & 0.82 \\
    ANN      & 0.94 & 0.94 & 0.94 & 0.94 \\
    \bottomrule
  \end{tabular}
   \caption{Performance Metrics of Deep Learning Models}
   \label{T6}
\end{table}

Both DBNs and DBMs achieved remarkable scores exceeding 0.94 on different evaluation metrics, as shown in \autoref{T7}.  
\begin{table}[htbp]
  \centering
  \small 
  
  \setlength{\tabcolsep}{4pt} 
  \begin{tabular}{p{0.17\columnwidth}cccc} 
    \toprule
    \textbf{Model Name} & \textbf{Precision} & \textbf{Recall} & \textbf{F1-Score} & \textbf{Accuracy} \\
    \midrule
    DBNs          & 0.95 & 0.94 & 0.94 & 0.94 \\
    DBMs       & 0.96 & 0.95 & 0.95 & 0.95 \\
    \bottomrule
  \end{tabular}
  \caption{Performance Metrics of RBMs Models}
  \label{T7}
\end{table}

\begin{figure}[t]
  \centering
  \includegraphics[width=0.9\textwidth]{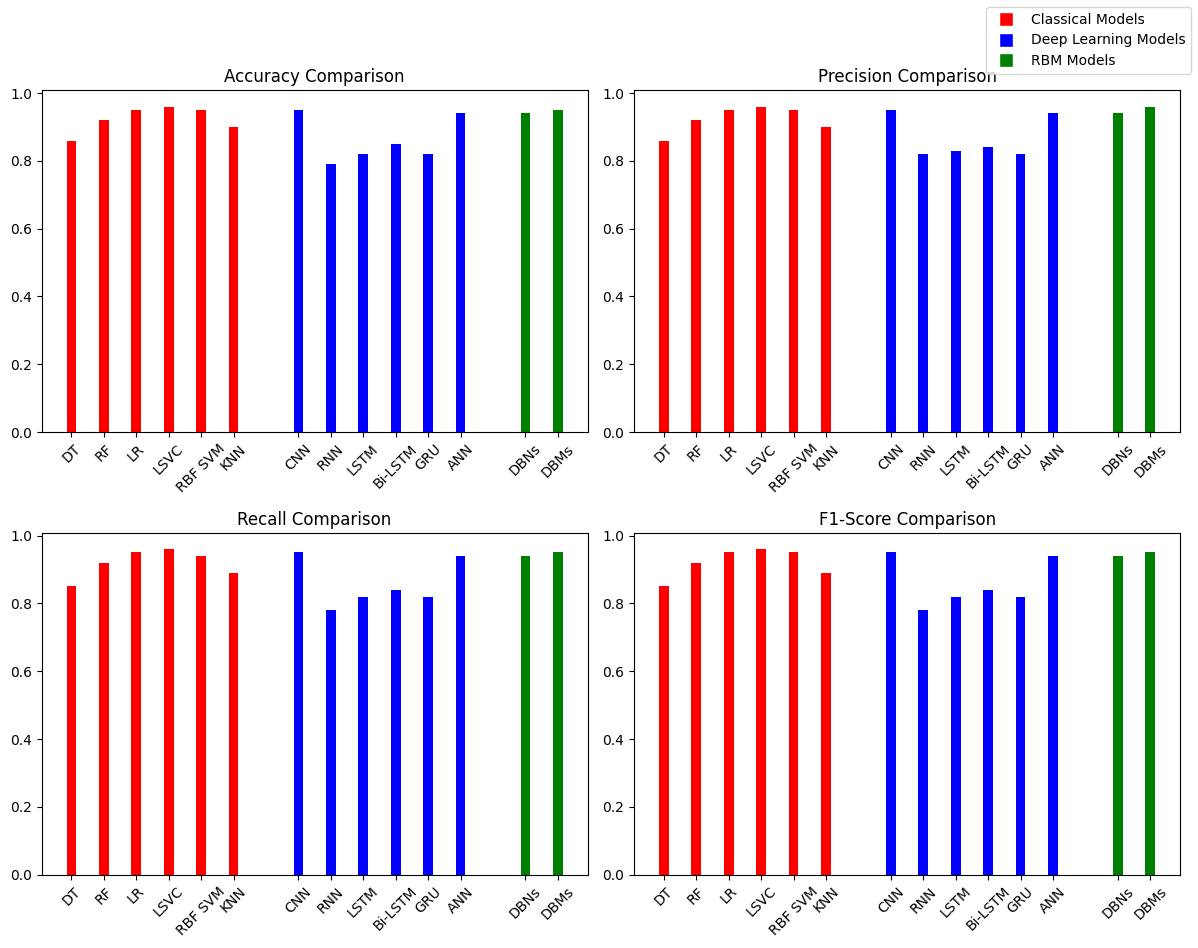}
  \caption{Performance Comparison Across Model Types}
  \label{F12}
\end{figure}

\begin{figure}[t]
  \centering
  \includegraphics[width=0.9\textwidth]{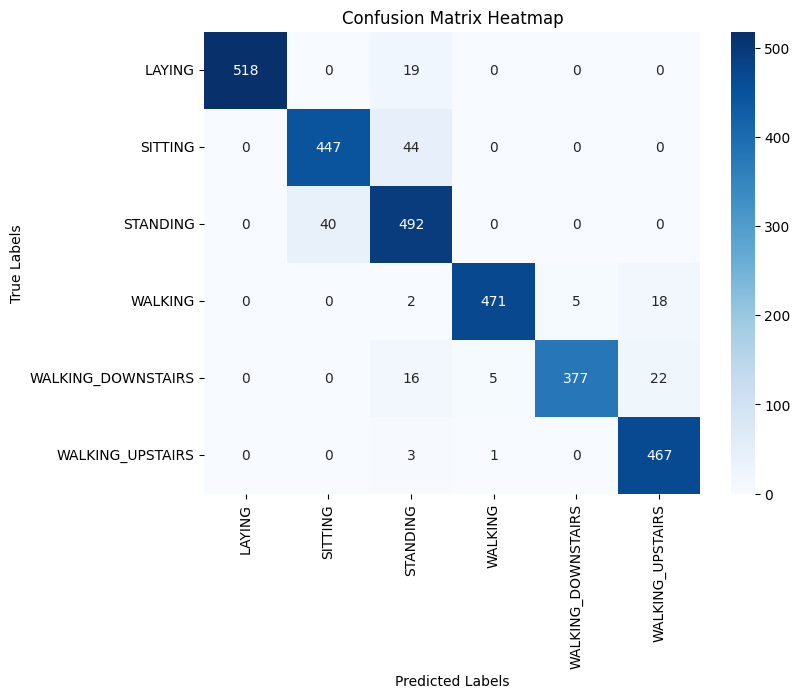}
  \caption{Confusion matrix for DMBs }
  \label{F13}
\end{figure}

\autoref{F12} provides a visual comparison of the performance metrics of classical Machine learning, deep learning and RBMs models on UCI-HAR dataset. Based on the average performance metrics, the RBM models have the highest average accuracy, precision, recall, and F1 score compared to classical machine learning and deep learning models. Within RBMs models, DBMs outperformed DBNs across all metrics. 
\autoref{F13} shows the confusion matrix obtained for the DBMs model. The DBM model performs at categorizing separate activities like "LAYING" and "WALKING," but struggles to distinguish between comparable postures like "SITTING" against "STANDING" and "WALKING-DOWNSTAIRS" against "WALKING-UPSTAIRS." This shows that additional feature extraction or model changes are required to better capture the small distinctions between these activities.

\subsection{Result for opportunity dataset}
525,660 samples were in the opportunity dataset, 341, 679 were used in training, and 183,981 were used in testing process. After training and testing each model type, the experiment results indicated that random forest achieved exception performance with accuracy (0.92), precision (0.94), recall (0.93), and F1-score (0.93), see \autoref{T8}.
\begin{table}[htbp]
  \centering
  \small 
  \setlength{\tabcolsep}{4pt} 
  \begin{tabular}{p{0.17\columnwidth}cccc} 
    \toprule
    \textbf{Model Name} & \textbf{Precision} & \textbf{Recall} & \textbf{F1-Score} & \textbf{Accuracy} \\
    \midrule
    DT         & 0.88 & 0.88 & 0.88 & 0.86 \\
   RF         & 0.94 & 0.93 & 0.93 & 0.92 \\
    LR  & 0.66 & 0.65 & 0.62 & 0.62 \\
    Linear SVC            & 0.37 & 0.32 & 0.32 & 0.31 \\
    RBF SVM               & 0.80 & 0.78 & 0.77 & 0.76 \\
    KNN  & 0.93 & 0.93 & 0.93 & 0.92 \\
    \bottomrule
  \end{tabular}
   \caption{Performance Metrics of classical Models}
   \label{T8}
\end{table}

Decision tree and K-nearest neighbour achieved impressive results. Linear Regression Linear SVC and RBF SVM did not perform well due to limitations in handling the complexity of big data.  

The result in \autoref{T9}  shows that in the deep learning model, CNN achieved the highest scores in performance metrics with impressive accuracy (0.89), precision (0.91), recall (0.90) and F1-score (0.90). Table indicates that Artificial neural network (ANN) and Bi-directional Long Short-term Memory (LSTM) are still showing strong performance, while Recurrent Neural Network (RNN), Long Short-term Memory (LSTM), and Gated Recurrent Units (GRU) additional optimization not having better performance compare with other models. 
\begin{table}[htbp]
  \centering
  \small 
  \setlength{\tabcolsep}{4pt} 
  \begin{tabular}{p{0.17\columnwidth}cccc} 
    \toprule
    \textbf{Model Name} & \textbf{Precision} & \textbf{Recall} & \textbf{F1-Score} & \textbf{Accuracy} \\
    \midrule
    CNN                     & 0.91 & 0.90 & 0.90 & 0.89 \\
    RNN                        & 0.61 & 0.68 & 0.60 & 0.63 \\
    LSTM                          & 0.71 & 0.71 & 0.69 & 0.69 \\
    Bi-LSTM & 0.81 & 0.81 & 0.80 & 0.79 \\
    GRU                          & 0.69 & 0.74 & 0.70 & 0.70 \\
    ANN                & 0.90 & 0.87 & 0.88 & 0.86 \\
    \bottomrule
  \end{tabular}
  \caption{Performance Metrics of Deep Learning Models}
  \label{T9}
\end{table}

Restricted Boltzmann Machines (RBMs) delivered a remarkable performance in opportunity dataset with Deep Belief Networks (DBNs) and Deep Boltzmann Machines (DBMs) both models consistently achieved  \noindent impressive accuracy, precision, recall, and F1-score, as shown in \autoref{T10}. 
\begin{table}[htbp]
  \centering
  \small 
  
  \setlength{\tabcolsep}{4pt} 
  \begin{tabular}{p{0.17\columnwidth}cccc} 
    \toprule
    \textbf{Model Name} & \textbf{Precision} & \textbf{Recall} & \textbf{F1-Score} & \textbf{Accuracy} \\
    \midrule
    DBNs                    & 0.93 & 0.91 & 0.92 & 0.91 \\
    DBMs                 & 0.87 & 0.83 & 0.83 & 0.81 \\
    \bottomrule
  \end{tabular}
  \caption{Performance Metrics of RBMs Models}
  \label{T10}
\end{table}

\begin{figure}[t]
  \centering
  \includegraphics[width=0.9\textwidth]{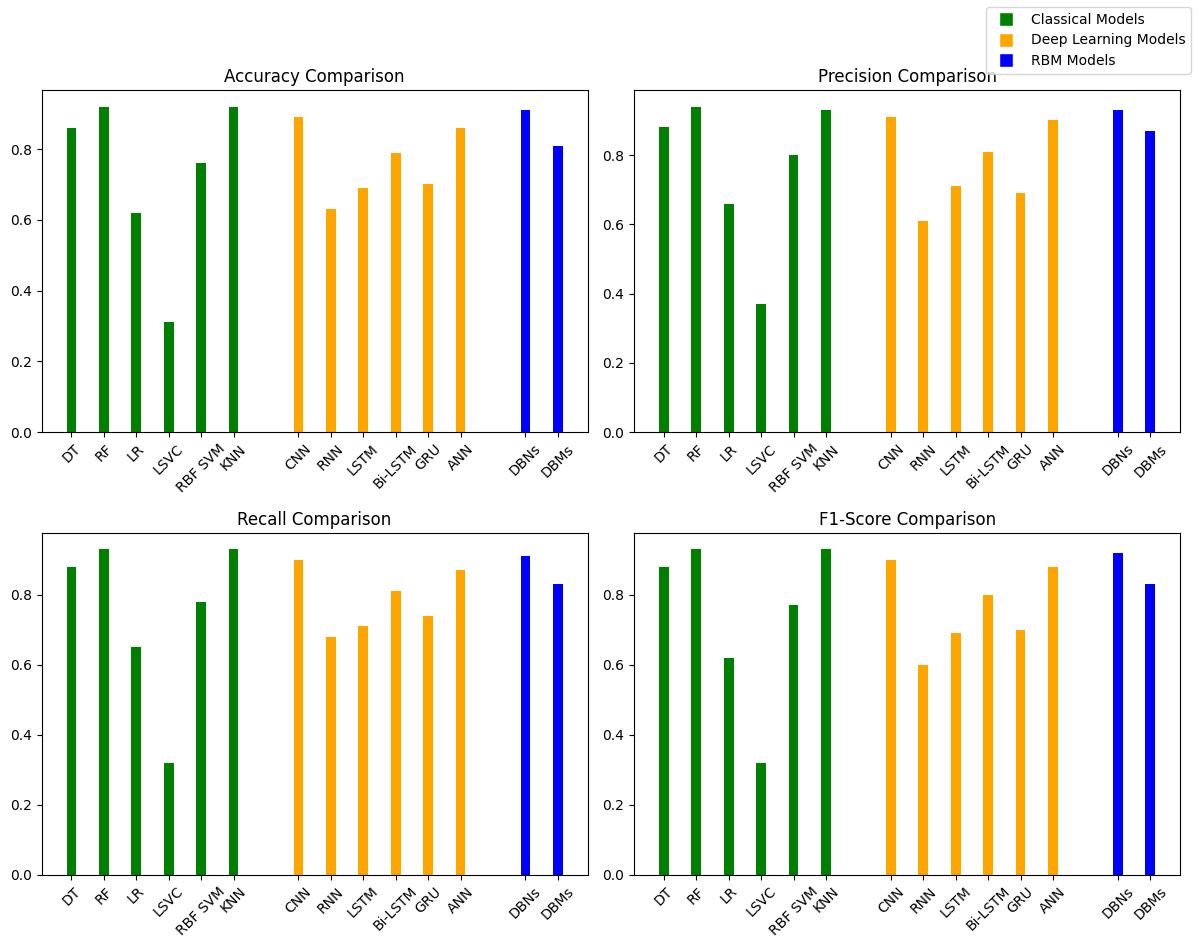}
  \caption{Performance Comparison Across Model Types.}
  \label{F14}
\end{figure}

\begin{figure}[H]
  \centering
  \includegraphics[width=0.8\textwidth]{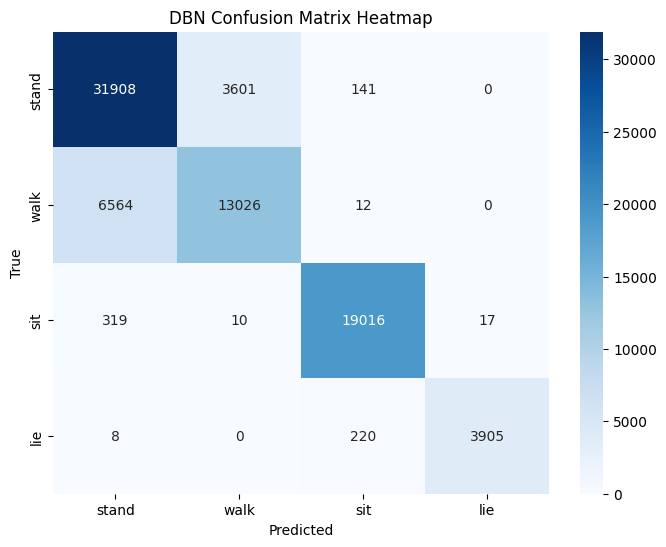}
  \caption{Confusion matrix for DBNs}
  \label{F15}
\end{figure}

\autoref{F14} shows the graphical representation of model performances on the opportunity dataset. Based on the average of performance metrics, Decision Tree within classical models, CNN within deep learning models and DBNs within Restricted Boltzmann Machines (RBMs) achieved the highest score.  However, among all the models DBNS within RBMs becomes the best performing model on the opportunity dataset. \autoref{F15}  depicts the confusion matrix of the DBNs model. Similar to DBMS model UCI-HAR dataset, DBNs model performs well for "stand" and "sit" activities, but there is significant confusion between "walk" and "stand," implying that the model's features may not adequately capture the differences between these activities, resulting in additional feature engineering or model adjustments to reduce this ambiguity. 

\subsection{Result for PAMAP2 (Physical Activity Monitoring Dataset)}

\noindent PAMAP2 dataset consists of 19,42,874 samples. Splitting the dataset into training and testing sets with 15,54,297 samples in the training dataset and 388,575 samples in the testing dataset. After training and testing, the result shows that Decision tree, Random Forest and k-nearest Neighbour in the classical model category achieved a performance metrics score of 1.0. Logistic regression, Linear SVC, and RBF SVM gained an impressive score in the evaluation, see  \autoref{T11}. 
\begin{table}[htbp]
  \centering
  \small 

  \setlength{\tabcolsep}{4pt} 
  \begin{tabular}{p{0.17\columnwidth}cccc} 
    \toprule
    \textbf{Model Name} & \textbf{Precision} & \textbf{Recall} & \textbf{F1-Score} & \textbf{Accuracy} \\
    \midrule
    DT        & 1.00              & 1.00            & 1.00               & 1.00             \\
    RF      & 1.00              & 1.00            & 1.00               & 1.00             \\
    LR & 0.91              & 0.90            & 0.90               & 0.93             \\
    Linear SVC           & 0.84              & 1.00            & 1.00               & 1.00             \\
    RBF SVM              & 0.99              & 0.99            & 0.99               & 0.99             \\
    KNN   & 1.00              & 1.00            & 1.00               & 1.00             \\
    \bottomrule
  \end{tabular}
  \caption{Performance Metrics of Classical Models}
  \label{T11}
\end{table}

In the deep learning category, CNN, ANN, and LSTM achieved performance metrics with scores ranging from 0.98 to 0.99, see \autoref{T12}. RBMs models,  DBNs and DBMs also performed remarkably well with impressive scores ranging from 0.97 to 0.99 across all metrics, see \autoref{T13}. 
\begin{table}[htbp]
  \centering
  \small 
  \setlength{\tabcolsep}{4pt} 
  \begin{tabular}{p{0.17\columnwidth}cccc} 
    \toprule
    \textbf{Model Name} & \textbf{Precision} & \textbf{Recall} & \textbf{F1-Score} & \textbf{Accuracy} \\
    \midrule
    CNN         & 0.99 & 0.99 & 0.99 & 0.99 \\
    RNN            & 0.85 & 0.84 & 0.84 & 0.87 \\
    LSTM              & 0.98 & 0.98 & 0.98 & 0.98 \\
    Bi-LSTM & 0.99 & 0.99 & 0.99 & 0.99 \\
    GRU               & 0.98 & 0.98 & 0.98 & 0.98 \\
    ANN      & 0.98 & 0.98 & 0.98 & 0.98 \\
    \bottomrule
  \end{tabular}
  \caption{Performance Metrics of Deep Learning Models}
  \label{T12}
\end{table}
\begin{table}[htbp]
  \centering
  \small 
  \setlength{\tabcolsep}{4pt} 
  \begin{tabular}{p{0.17\columnwidth}cccc} 
    \toprule
    \textbf{Model Name} & \textbf{Precision} & \textbf{Recall} & \textbf{F1-Score} & \textbf{Accuracy} \\
    \midrule
     DBNs         & 0.97 & 0.97 & 0.97 & 0.98 \\
     DBMs     & 0.98 & 0.98 & 0.98 & 0.99 \\
    \bottomrule
  \end{tabular}
    \caption{Performance Metrics of RBMs Models}
    \label{T13}
\end{table}

\begin{figure}[t]
  \centering\includegraphics[width=0.9\textwidth]{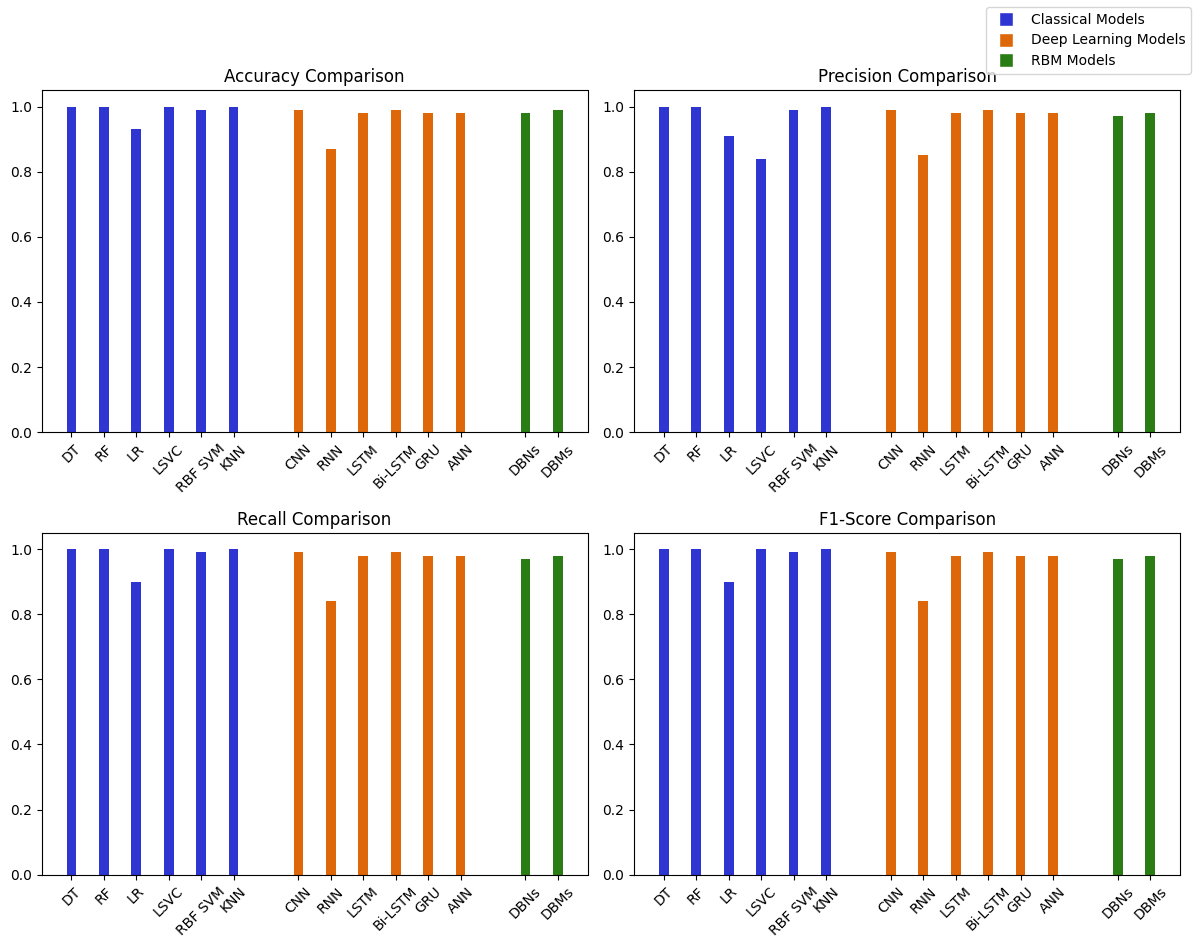}\caption{Performance Comparison Across Model Types.}
  \label{F16}
\end{figure}

Based on the overall result shown in  \autoref{F16},  classical machine learning and deep learning models perform well in classifying human activity using PAMAP2 datasets.
Among all the model types deep learning models are considered to be the highest performers in this dataset, particularly CNN and BiLSTM. Although classical models achieved perfect scores of 1.0 across all metrics in several models.
Perfect scores on training data do not guarantee the same performance on unseen data. There is a risk of overfitting, where the model learns to memorize the training data instead of capturing patterns that generalize well to new or unseen data. Deep learning models, with their ability to learn the hierarchical representation of data, often generalize better to unseen data. 

Moreover, RBMs based model also performed well, as it is a better option for tasks involving feature learning and representation. Due to all these reasons, CNN model become the best-performing model on PAMAP2 dataset, see  \autoref{F17}.  
\begin{figure}[H]
  \centering
  \includegraphics[width=0.9\textwidth]{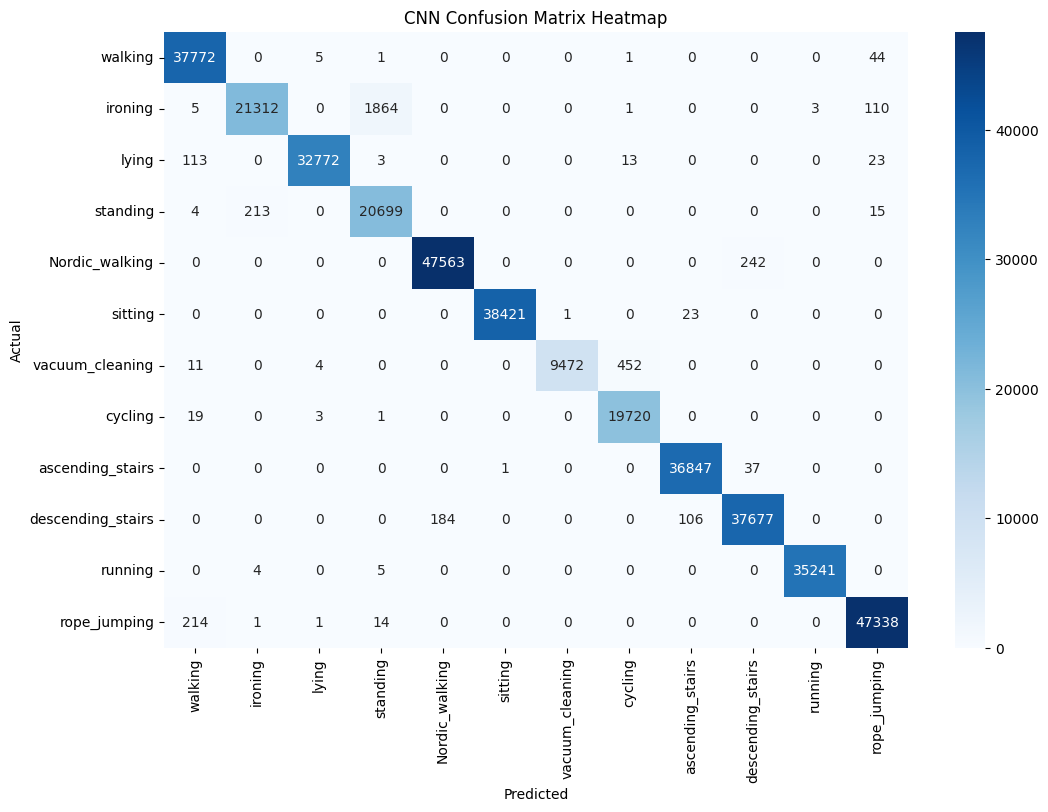}
  \caption{Confusion matrix for CNN}
\end{figure}
\label{F17}
\subsection{Result for WISDM (Wireless Sensor Data Mining Dataset) }

WISDM dataset consists of 1,073,623 samples with 858,898 samples used for training and 214,725 for testing. Among classical model, Random Forest achieved the highest performance across with scores of accuracy (0.64), precision (0.63), recall (0.60) and F1-score (0.60), as shown in \autoref{T14}. Decision Tree also showed promising results with a F1-Score of 0.56, although it had lower precision, recall and accuracy compared to random forest. 
\begin{table}[htbp]
  \centering
  \small 
 
  \label{tab:performance66}
  \setlength{\tabcolsep}{4pt} 
  \begin{tabular}{p{0.17\columnwidth}cccc} 
    \toprule
    \textbf{Model Name} & \textbf{Precision} & \textbf{Recall} & \textbf{F1-Score} & \textbf{Accuracy} \\
    \midrule
    DT       & 0.56               & 0.56            & 0.56              & 0.54 \\
    RF       & 0.63               & 0.60            & 0.60              & 0.64 \\
    LR  & 0.25               & 0.29            & 0.26              & 0.48 \\
    Linear SVC           & 0.23               & 0.24            & 0.22              & 0.47 \\
    RBF SVM              & 0.33               & 0.38            & 0.35              & 0.57 \\
    KNN  & 0.57               & 0.58            & 0.57              & 0.58 \\
    \bottomrule
  \end{tabular}
   \caption{Performance Metrics of Classical Models}
   \label{T14}
\end{table}
The other classical models including Logistic regression, Linear SVC, RBF SVM and KNN demonstrated lower scores across all the metrics (Table).  

In deep learning, CNN, RNN, LSTM, BiLSTM and GRU showed similar performance with F1-score ranging from 0.53 to 0.55 and accuracies around 0.63. ANN had lower results compared to other models, see \autoref{T15}.
\begin{table}[htbp]
  \centering
  \small 
  
  \setlength{\tabcolsep}{4pt} 
  \begin{tabular}{p{0.17\columnwidth}cccc} 
    \toprule
    \textbf{Model Name} & \textbf{Precision} & \textbf{Recall} & \textbf{F1-Score} & \textbf{Accuracy} \\
    \midrule
    CNN             & 0.62             & 0.57            & 0.55              & 0.63 \\
    RNN                & 0.63             & 0.56            & 0.53              & 0.63 \\
    LSTM                   & 0.61             & 0.57            & 0.54              & 0.63 \\
    Bi-LSTM   & 0.61             & 0.57            & 0.54              & 0.63 \\
    GRU                    & 0.62             & 0.57            & 0.53              & 0.63 \\
    ANN          & 0.58             & 0.57            & 0.52              & 0.52 \\
    \bottomrule
  \end{tabular}
   \caption{Performance Metrics of Deep Learning Models}
   \label{T15}
\end{table}

\begin{figure}[t]
  \centering
  \includegraphics[width=0.9\textwidth]{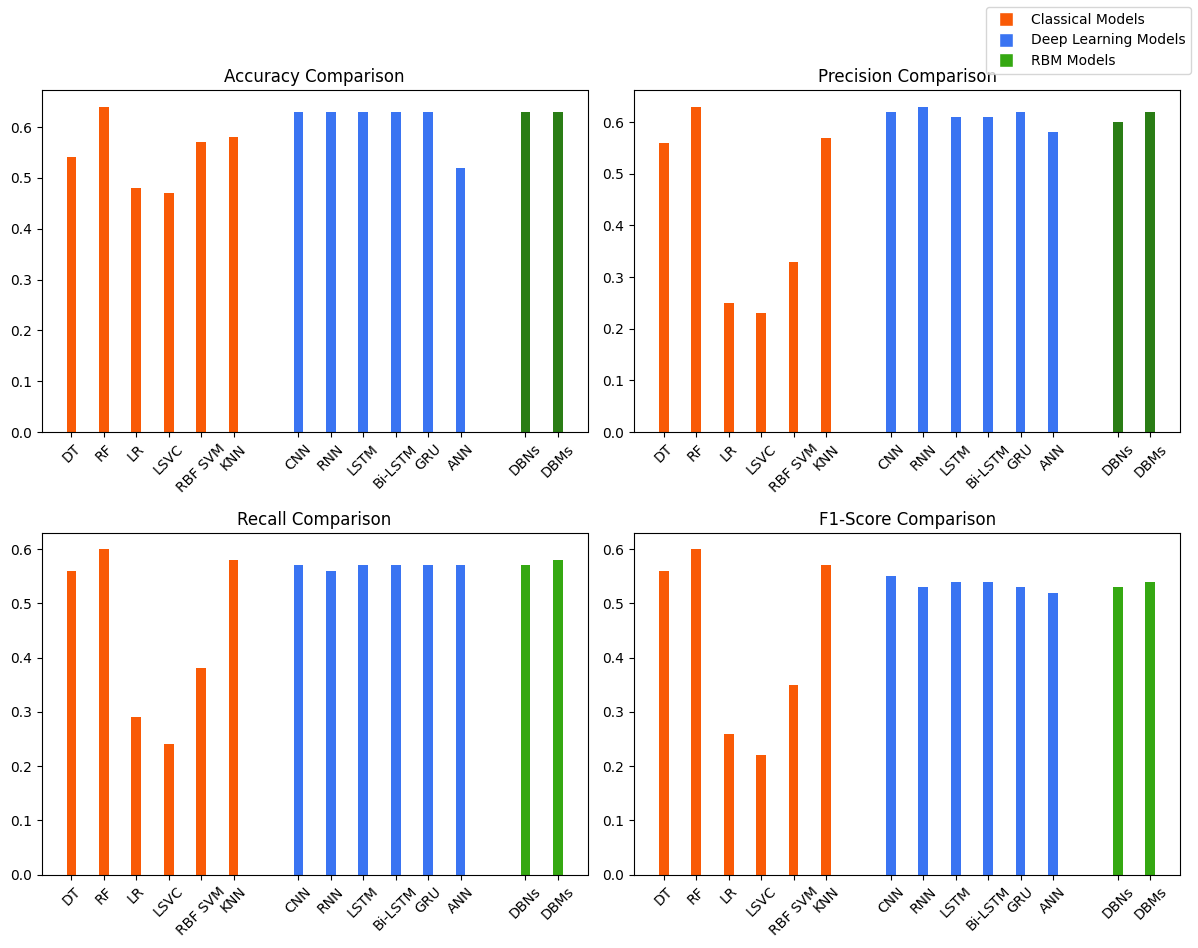}
  \caption{Performance Comparison Across Model Types.}
  \label{F18}
\end{figure}

\begin{figure}[H]
  \centering
  \includegraphics[width=0.9\textwidth]{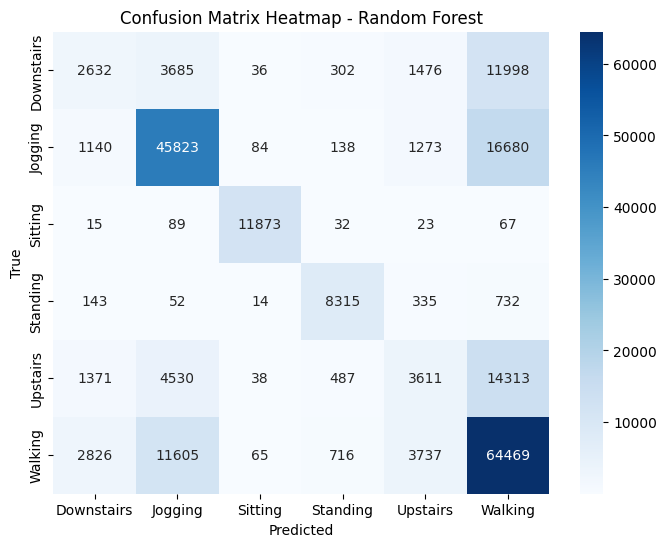}
  \caption{Confusion matrix of Random Forest}
  \label{F19}
\end{figure}

 DBNs and DBMs in the RBMs-based models showed consistent performance with F1-score around 0.53 to 0.54 and accuracy 0.63, see  \autoref{T16}.  
\begin{table}[htbp]
  \centering
  \small 
  \setlength{\tabcolsep}{4pt} 
  \begin{tabular}{p{0.17\columnwidth}cccc} 
    \toprule
    \textbf{Model Name} & \textbf{Precision} & \textbf{Recall} & \textbf{F1-Score} & \textbf{Accuracy} \\
    \midrule
    DBNs             & 0.60             & 0.57            & 0.53              & 0.63 \\
    DBMs           & 0.62             & 0.58            & 0.54              & 0.63 \\
    \bottomrule
  \end{tabular}
   \caption{Performance Metrics of RBMs Models}
   \label{T16}
\end{table}
\autoref{F18} shows the overall comparison of all the models, Random Forest.  \autoref{F19} achieved the performance metrics compared to other types of models. Logistic regression, Linear SVC, RBF SVM and KNN show significant imbalances in performance across different activity classes. These models failed to make accurate predictions for 'Downstairs', 'Standing', and 'Upstairs' which caused them to lower performance scores. All the deep learning models achieved the same accuracy score of 0.63. WISDM dataset lacks variability or contains biases that may limit the ability of these models to learn distinct patterns resulting in similar overall performance. RBM approaches showed competitive performance comparable to deep learning methods in handling WISDM tasks. 
\subsection{Result for Berkeley MHAD Dataset (Multimodal Human Action Database)     }

The Berkely dataset consists of a large number of samples totaling 2, 401, 920 instances with 1,801,440 samples in the training dataset and 600,480 samples in the testing dataset. The training set is used to train the models on the labeled data allowing them to learn the patterns and relationship with the dataset. The testing dataset, which is unseen during training, is used to evaluate the performance of the trained models on new and unseen data. In the evaluation of classical models, Decision trees, Random Forest, and Logistic Regression achieved near-perfect performance metrics score consistently at or above 0.99. However, Linear SVC and RBF SVM showed significantly lower performance \autoref{T17} score, proving that these models might not be well suited for this dataset.
\begin{table}[htbp]
  \centering
  \small 
 
  \setlength{\tabcolsep}{4pt} 
  \begin{tabular}{p{0.17\columnwidth}cccc} 
    \toprule
    \textbf{Model Name} & \textbf{Precision} & \textbf{Recall} & \textbf{F1-Score} & \textbf{Accuracy} \\
    \midrule
    Decision Tree       & 0.99               & 0.99            & 0.99              & 1.00 \\
    Random Forest       & 1.00               & 1.00            & 1.00              & 1.00 \\
    LR & 1.00               & 1.00            & 1.00              & 1.00 \\
    Linear SVC          & 0.23               & 0.20            & 0.17              & 0.23 \\
    RBF SVM             & 0.93               & 0.08            & 0.03              & 0.21 \\
    KNN & 1.00               & 1.00            & 1.00              & 1.00 \\
    \bottomrule
  \end{tabular}
   \caption{Performance Metrics of Classical Models}
   \label{T17}
\end{table}

In deep learning, CNN achieved impressive performance with an accuracy of approximately 0.96, 0.94, 0.95 and 0.97 respectively. RNNs and variants such as LSTM, BiLSTM, and GRU among them LSTM and BiLSTM outperformed others \autoref{T18}. 
\begin{table}[htbp]
  \centering
  \small 
  \setlength{\tabcolsep}{4pt} 
  \begin{tabular}{p{0.17\columnwidth}cccc} 
    \toprule
    \textbf{Model Name} & \textbf{Precision} & \textbf{Recall} & \textbf{F1-Score} & \textbf{Accuracy} \\
    \midrule
    CNN           & 0.96             & 0.94            & 0.95               & 0.97 \\
    RNN                & 0.62             & 0.40            & 0.40               & 0.59 \\
    LSTM                  & 0.93             & 0.91            & 0.92               & 0.95 \\
    Bi-LSTM   & 0.94             & 0.95            & 0.94               & 0.96 \\
    GRU                   & 0.91             & 0.87            & 0.88               & 0.92 \\
    ANN         & 0.86             & 0.81            & 0.82               & 0.89 \\
    \bottomrule
  \end{tabular}
   \caption{Performance Metrics of Deep Learning Models}
   \label{T18}
\end{table}

The RBMs models, DBNs and DBMs, demonstrated promising results, Both DBNs and DBMs achieved high performance metrics score \autoref{T19} indicating their effectiveness in capturing patterns and dependencies. 
\begin{table}[htbp]
  \centering
  \small 
  \setlength{\tabcolsep}{4pt} 
  \begin{tabular}{p{0.17\columnwidth}cccc} 
    \toprule
    \textbf{Model Name} & \textbf{Precision} & \textbf{Recall} & \textbf{F1-Score} & \textbf{Accuracy} \\
    \midrule
    DBNs            & 0.86             & 0.79            & 0.79               & 0.88 \\
    DBMs         & 0.93             & 0.90            & 0.91               & 0.94 \\
    \bottomrule
  \end{tabular}
  \caption{Performance Metrics of RBMs Models}
  \label{T19}
\end{table}
\begin{figure}[H]
  \centering
  \includegraphics[width=0.9\textwidth]{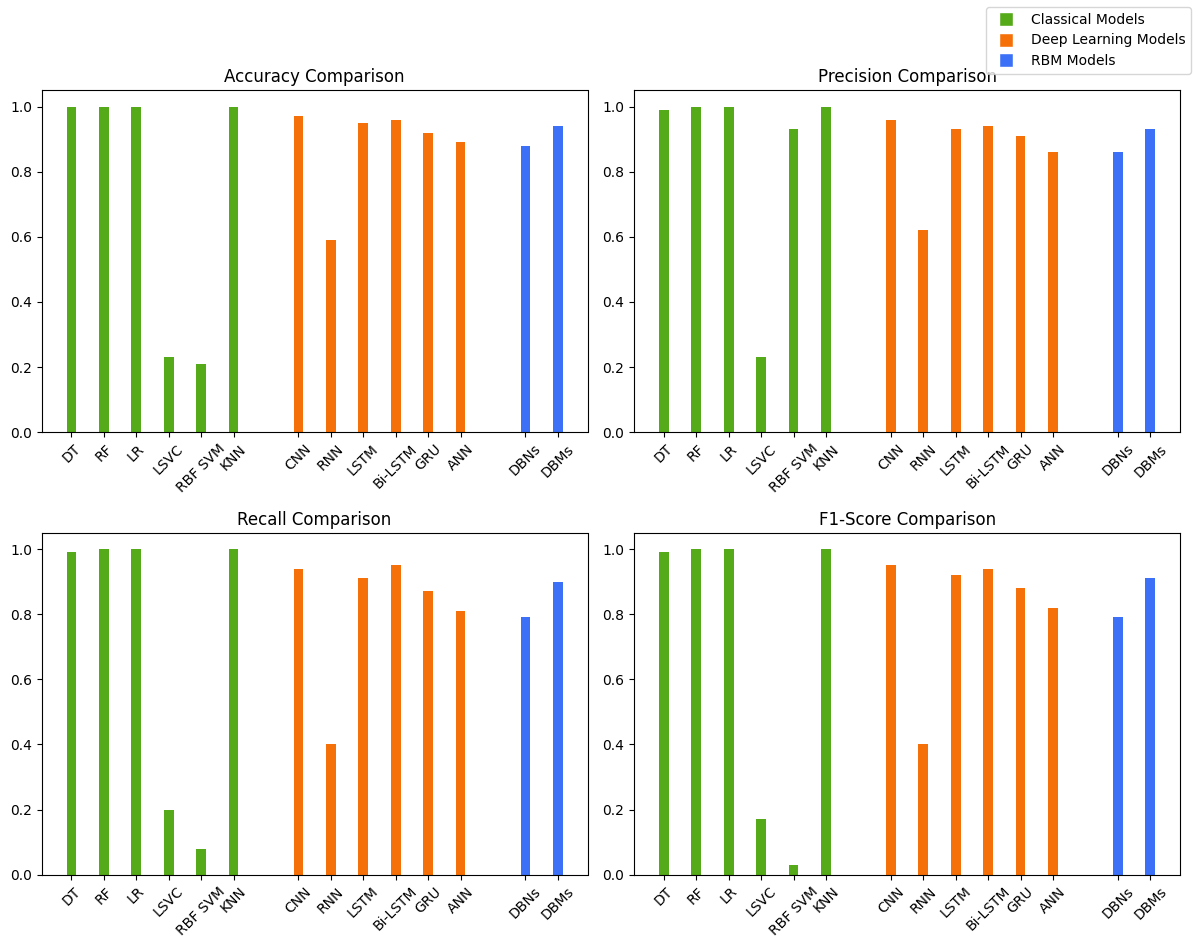}
  \caption{Performance Comparison Across Model Types.}
  \label{F20}
\end{figure}

\begin{figure}[H]
  \centering
  \includegraphics[width=0.9\textwidth]{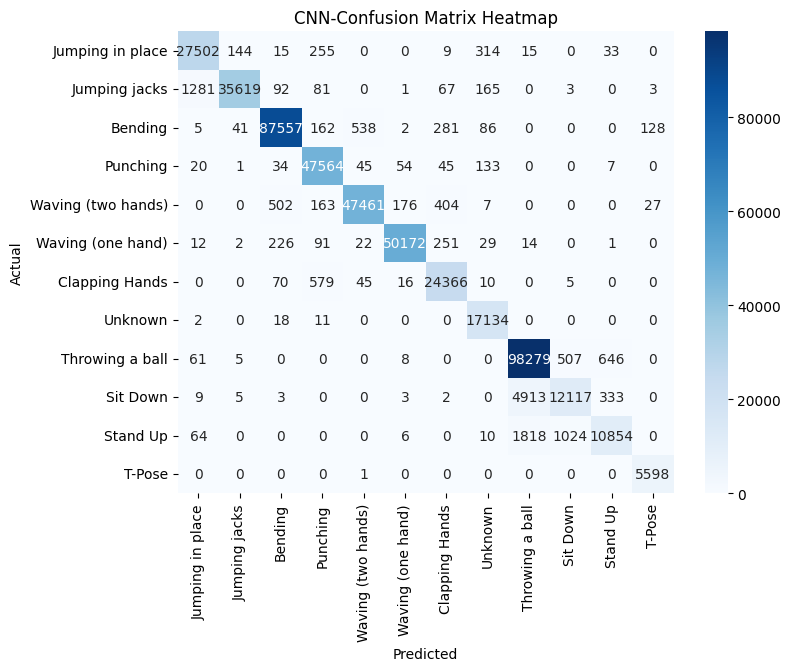}
  \caption{Confusion matrix for CNN}
  \label{F21}
\end{figure}

Overall, the comparison among three model types  reveals that deep learning models particularly CNN model offer excellent performance for human activity recognition tasks on the Berkeley MHAD dataset because they naturally learn and adapt to subtle movement patterns directly from the data (see \autoref{F20}). While classical models, such as Decision Trees, Random Forests, SVMs, and KNN, rely on predetermined features and may overlook these tiny subtleties, CNNs automatically capture and modify them, resulting in considerably higher accuracy  (see \autoref{F21}). 
CNNs offer a significant advantage due to their capability to discern complex variations between different activities.

\section{Limitations and Future Work}
While this study demonstrates promising results in human activity recognition using different machine learning model families, several limitations highlight opportunities for future research. The high dependency on labelled sensor data, which is frequently limited in diversity, limits model generalizability, and class imbalances result in biased predictions. Moreover, we experiment on sensor data alone, which, although effective, may not fully capture the complexity of human activities, particularly in more subtle or overlapping contexts.

Future research should explore hybrid approaches that combine classical and deep learning models, experimenting with ensemble methods for better performance and flexibility. Additionally, transfer learning could improve model generalization across diverse datasets, reducing dependency on large labelled data. Addressing imbalanced datasets through oversampling or synthetic data generation is crucial for fair performance. Integrating multi-modal data, such as audio or video inputs, could further enhance the accuracy and applicability of HAR systems, particularly in areas like healthcare and surveillance. Additionally, integrating explainable AI (XAI) into human activity recognition models improves transparency by revealing key features influencing model decisions and fostering trust and adoption in critical applications. Predominantly, the focus should be developing interpretable, generalizable, and robust HAR models to improve practical outcomes in various real-world applications.

Moreover, the accurate recognition of human activities, as explored in our comparative analysis of HAR models, forms a crucial foundation for higher-level cognitive tasks such as intention and plan recognition \citep{pereira2013state,sukthankar2014plan,HanBook2013,sadri2011logic,vellenga2024pt,di2023recognition}.  Understanding the what of an activity (e.g., walking, sitting) is only the first step;  inferring the why (e.g., going to a meeting, resting) requires integrating temporal context, environmental factors, and prior knowledge about the individual's goals and routines.  Future research should explore how the  activity recognition models presented here can be effectively integrated with higher-level  intention and plan recognition techniques to create more comprehensive and contextually aware systems.  This integration would allow for a deeper understanding of human behaviour, enabling more sophisticated applications in areas such as personalised healthcare, assistive robotics, and proactive security systems.

\section{Conclusion}

In this study, we significantly evaluated different types of machine learning models, including classical model, deep learning and Restricted Boltzmann Machines (RBMs), for human activity recognition tasks across four diverse datasets: UCI-HAR, OPPORTUNITY, PAMAP2, WISDM and Berkeley MHAD. Each dataset presented unique challenges and characteristics, offering a comprehensive experimentation for our analysis. Our findings indicate that the performance of the models varied significantly across datasets. Classical machine learning models particularly Random Forest, Decision Tree, and K-nearest neighbors demonstrated strong performance on smaller datasets like UCI-HAR and PAMAP2 achieving high accuracy, precision, recall and F1-score. However, their performance degraded on longer and more complex datasets like opportunity and WISDM, highlighting their limitations in handling big data and capturing intricate patterns. Deep learning models, especially CNN consistently outperformed classical models across all datasets, showing their capabilities to learn hierarchical representations. CNN performed remarkably on all datasets, particularly in the Berkeley MHAD dataset. RBMs specifically, DBNs and DBMs, showed promising results across all datasets indicating their potential for feature learning and representation. However, their performance is slightly lower than deep learning models suggesting a trade-off between interpretability and performance.
These findings have important practical implications for HAR system development. The improved performance of deep learning models implies that they are suitable for applications that need accurate and dependable activity recognition. In healthcare, for example, robust HAR systems can improve patient monitoring, allowing for quicker interventions and individualized care plans. In surveillance scenarios, these systems can enhance security measures by accurately evaluating human actions in various environments, improving incident detection and response. Overall, this study emphasizes the importance of advanced machine learning approaches in the growth of human activity recognition, paving the way for innovative applications in domains as diverse as healthcare, surveillance, and beyond. 

\bibliographystyle{unsrtnat}
\bibliography{references}
\setcitestyle{numbers,square}
\end{document}